\documentclass[10pt,twocolumn,letterpaper]{article}
\usepackage{cvpr}
\usepackage{times}
\usepackage{epsfig}
\usepackage{graphicx}
\usepackage{amsmath}
\usepackage{amssymb}
\usepackage{amsmath}
\usepackage{amssymb}
\usepackage{dsfont}
\usepackage{bm}
\usepackage{xcolor,soul,colortbl}
\usepackage{array}
\usepackage{multirow}
\usepackage{subcaption}
\usepackage{dashrule}

\newcommand{\THVT}[1]{\textcolor{black}{#1}}
\newcommand{\THVO}[1]{\textcolor{black}{#1}}

\definecolor{CA}{HTML}{E6E6FA}
\definecolor{CS}{HTML}{FFEFD5}
\definecolor{CE}{HTML}{FFC0CB}

\definecolor{tcolor}{RGB}{219,50,54}
\definecolor{scolor}{RGB}{72,133,237}

\newcommand{\mm}[1]{\ensuremath{\bm{#1}}} 
\newcommand{\rtb}[1]{\rotatebox{90}{#1}}   
\newcommand{\reff}[1]{(\ref{#1})}

\newcolumntype{C}[1]{>{\centering\arraybackslash}p{#1}}	
\newcolumntype{L}[1]{>{\raggedright\arraybackslash}p{#1}}
\newcolumntype{R}[1]{>{\raggedleft\arraybackslash}p{#1}}

\DeclareMathOperator{\E}{\mathbb{E}}

\newcommand{\src}[1]{#1_s}
\newcommand{\trg}[1]{#1_t}
\newcommand{\cX}{\mathcal{X}}
\newcommand{\cY}{\mathcal{Y}}
\newcommand{\bx}{\mm x}
\newcommand{\by}{\mm y}

\usepackage{dsfont}
\usepackage{flushend}
\usepackage[utf8]{inputenc}
\usepackage{authblk}

\usepackage[pagebackref=true,breaklinks=true,letterpaper=true,colorlinks,bookmarks=false]{hyperref}

\cvprfinalcopy 

\def\cvprPaperID{396} 

\begin{document}
	
	\title{ADVENT: Adversarial Entropy Minimization for Domain Adaptation \\in Semantic Segmentation}
	\makeatletter
	\renewcommand\AB@affilsepx{\qquad\qquad \protect\Affilfont}
	\makeatother
	\author[1]{Tuan-Hung Vu}
	\author[1]{Himalaya Jain}
	\author[1]{Maxime Bucher}
	\author[1,2]{Matthieu Cord}
	\author[1]{Patrick Pérez}
	\affil[1]{valeo.ai, Paris, France}
	\affil[2]{Sorbonne University, Paris, France}
	
	\maketitle
	\thispagestyle{empty}
	
	\begin{abstract}
		Semantic segmentation is a key problem for many computer vision tasks.
		While approaches based on convolutional neural networks constantly break new records on different benchmarks, generalizing well to diverse testing environments remains a major challenge.
		In numerous real world applications, there is indeed a large gap between data distributions in train and test domains, which results in severe performance loss at run-time.
		In this work, we address the task of unsupervised domain adaptation in semantic segmentation with losses based on the entropy of the pixel-wise predictions. 
		To this end, we propose two novel, complementary methods using (i) an entropy loss and (ii) an adversarial loss respectively.
		We demonstrate state-of-the-art performance in semantic segmentation on two challenging ``synthetic-2-real'' set-ups\footnote{Code available at \url{https://github.com/valeoai/ADVENT}.} 
		and show that the approach can also be used for detection.
	\end{abstract}
	
	\section{Introduction}
	Semantic segmentation is the task of assigning class labels to all pixels in an image.
In practice, segmentation models often serve as the backbone in complex computer vision systems like autonomous vehicles, which demand high accuracy in a large variety of urban environments.
For example, under adverse weathers, the system must be able to recognize roads, lanes, sideways or pedestrians despite their appearances being largely different from ones in the training set.
A more extreme and important example is so-called ``synthetic-2-real'' set-up~\cite{Ros_2016_CVPR,Richter_2016_ECCV} -- training samples are synthesized by game engines and test samples are real scenes.
Current fully-supervised approaches~\cite{long2015fully,Zhao_2017_CVPR,chen2018deeplab} have not yet guaranteed a good generalization to arbitrary test cases.
Thus a model trained on one domain, named as \textit{source}, usually undergoes a drastic drop in performance when applied on another domain, named as \textit{target}.

Unsupervised domain adaptation (UDA) is the field of research that aims at learning only from source supervision a well performing model on target samples.
Among the recent methods for UDA, many address the problem by reducing cross-domain discrepancy, along with the supervised training on the source domain.
They approach UDA by minimizing the difference between the distributions of the intermediate features or of the final outputs for source and target data respectively.
It is done at single \cite{hoffman2016fcns,saito2017adversarial,yan2017mind} or multiple levels \cite{long2015learning,long2016unsupervised} using maximum mean discrepancies (MMD) or adversarial training \cite{ganin2015unsupervised,tzeng2017adversarial}.
Other approaches include self-training \cite{zou2018unsupervised} to provide pseudo labels or generative networks to produce target data \cite{hoffman18a, Sankaranarayanan_2018_CVPR, wu2018dcan}.

\begin{figure}[t!]
	\centering
	\includegraphics[width=0.47\textwidth]{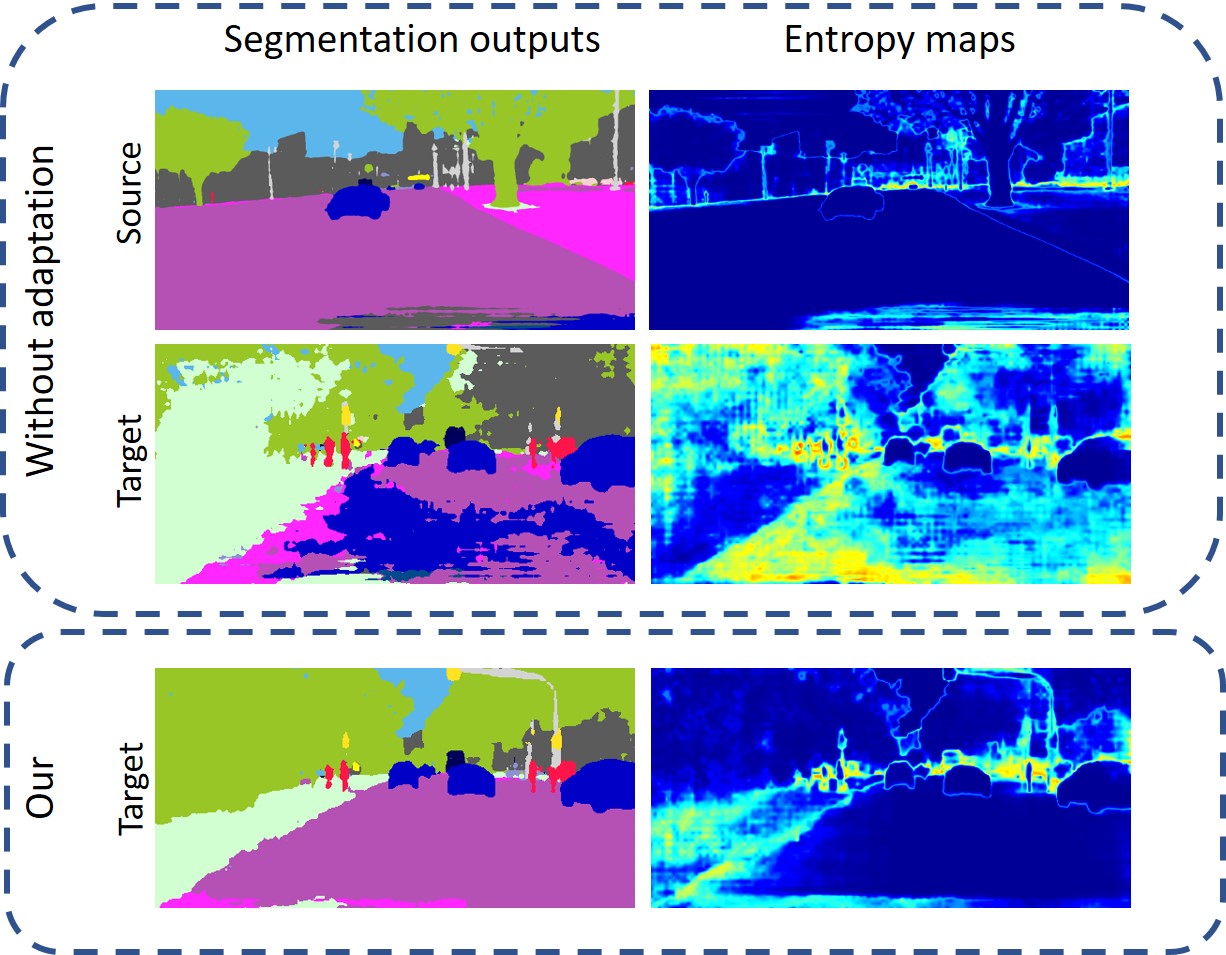}
	\vspace{-0.2cm}\caption{\small \textbf{Proposed entropy-based unsupervised domain adaptation for semantic segmentation}. The top two rows show results on source and target domain scenes of the model trained without adaptation.  The bottom row shows the result on the same target domain scene of the model trained with entropy-based adaptation. The left and right columns visualize respectively the semantic segmentation outputs and the corresponding prediction entropy maps (see text for details).}
	\label{lbl:teaser}
\end{figure}

Semi-supervised learning addresses a closely related problem of learning from the data of which only a subset is annotated. Thus, it inspires several approaches for UDA, for example, self-training, generative model or class balancing~\cite{zhu05survey}.
\textit{Entropy minimization} is also one of the successful approaches used for semi-supervised learning \cite{springenberg2015unsupervised}.

In this work, we adapt the principle of entropy minimization to the UDA task in semantic segmentation.
We start from a simple observation: models trained only on source domain tend to produce \textit{over-confident}, \ie, low-entropy, predictions on source-like images and \textit{under-confident}, \ie, high-entropy, predictions on target-like ones.
Such a phenomenon is illustrated in Figure~\ref{lbl:teaser}.
Prediction entropy maps of scenes from the source domain look like edge detection results with high entropy activations only along object borders.
On the other hand, predictions on target images are less certain, resulting in very noisy, high entropy outputs.
We argue that one possible way to bridge the domain gap between source and target is by enforcing high prediction certainty (low-entropy) on target predictions as well.
To this end, we propose two approaches:  direct entropy minimization using an entropy loss and indirect entropy minimization using an adversarial loss.
While the first approach imposes the low-entropy constraint on independent pixel-wise predictions, the latter aims at globally matching source and target distributions in terms of \textit{weighted self-information}.\footnote{Connection to the entropy is discussed in Section~\ref{sec:approach}.} 
We summarize our contributions as follows:

\begin{itemize}
	\setlength{\parskip}{0pt}
	\setlength{\itemsep}{1pt}
	\item For semantic segmentation UDA, we propose to leverage an entropy loss to directly penalize low-confident predictions on target domain.
	The use of this entropy loss adds no significant overhead to existing semantic segmentation frameworks.
	\item We introduce a novel entropy-based adversarial training approach targeting not only the entropy minimization objective but also the structure adaptation from source domain to target domain.
	\item To improve further the performance in specific settings, we suggest two additional practices: (i) training on specific entropy ranges and (ii) incorporating class-ratio priors. We discuss practical insights in the experiments and ablation studies.
\end{itemize}
The entropy minimization objectives push the model's decision boundaries toward low-density regions of the target domain distribution in prediction space.
This results in ``cleaner'' semantic segmentation outputs, with more refined object edges as well as large ambiguous image regions being correctly recovered, as shown in Figure~\ref{lbl:teaser}.
The proposed models outperform state-of-the-art approaches on several UDA benchmarks for semantic segmentation, in particular the two main synthetic-2-real benchmarks, GTA5$\rightarrow$Cityscapes and SYNTHIA$\rightarrow$Cityscapes.
	
	\section{Related works}
	Unsupervised Domain Adaptation is a well researched topic for the task of classification and detection, with recent advances in semantic segmentation also. A very appealing application of domain adaptation is on using synthetic data for real world tasks. This has encouraged the development of several synthetic scene projects with associated datasets, such as Carla \cite{carla_Dosovitskiy17}, SYNTHIA \cite{Ros_2016_CVPR}, and others \cite{shafaei2016play, Richter_2016_ECCV}.

The main approaches for UDA include discrepancy minimization between source and target feature distributions \cite{ganin2015unsupervised,long2015learning,hoffman2016fcns,long2016unsupervised,tzeng2017adversarial}, self-training with pseudo-labels \cite{zou2018unsupervised} and generative approaches \cite{hoffman18a, Sankaranarayanan_2018_CVPR, wu2018dcan}.
In this work, we are particularly interested in UDA for the task of semantic segmentation. Therefore, we only review the UDA approaches for semantic segmentation here (see \cite{csurka2017domain} for a more general literature review). 

\textit{Adversarial training} for UDA is the most explored approach for semantic segmentation. It involves two networks. One network predicts the segmentation maps for the input image, which could be from source or target domain, while another network acts as a discriminator which takes the feature maps from the segmentation network and tries to predict domain of the input. The segmentation network tries to fool the discriminator, thus making the features from the two domains have a similar distribution.
Hoffman \etal \cite{hoffman2016fcns} are the first to apply the adversarial approach for UDA on semantic segmentation. They also have a category specific adaptation by transferring the label statistics from the source domain. A similar approach of global and class-wise alignment is used in \cite{chen2017no} with the class-wise alignment being done using adversarial training on grid-wise soft pseudo-labels. In \cite{chen2018CVPR}, adversarial training is used for spatial-aware adaptation along with a distillation loss to specifically address synthetic-2-real domain shift. \cite{hong2018CVPR} uses a residual net to make the source feature maps similar to target's ones using adversarial training, the feature maps being then used for the segmentation task.
In \cite{tsai2018learning}, the adversarial approach is used on the output space to benefit from the structural consistency across domain.
\cite{saito2017adversarial, saito2017maximum} propose another interesting way of using adversarial training: They get two predictions on the target domain image, this is done either by two classifiers \cite{saito2017maximum} or using dropout in the classifier \cite{saito2017adversarial}. Given the two predictions the classifier is trained to maximize the discrepancy between the distributions while the feature extractor part of the network is trained to minimize it.

Some methods build on \textit{generative networks} to generate target images conditioned on the source. Hoffman \etal \cite{hoffman18a} propose Cycle-Consistent Adversarial Domain Adaptation (CyCADA), in which they adapt at both pixel-level and feature-level representation. For pixel-level adaptation they use Cycle-GAN \cite{zhu2017unpaired} to generate target images conditioned on the source images. In \cite{Sankaranarayanan_2018_CVPR}, a generative model is learned to reconstruct images from the feature space. Then, for domain adaptation, the feature module is trained to produce target images on source features and vice-versa using the generator module. In DCAN \cite{wu2018dcan}, channel-wise feature alignment is used in the generator and segmentation network. The segmentation network is learned on generated images with the content of the source and style of the target for which source segmentation map serves as the ground-truth. The authors in \cite{zhu2018ECCV} use generative adversarial networks (GAN)~\cite{goodfellow2014generative} to align the source and target embeddings. Also, they replace the cross-entropy loss by a \textit{conservative loss} (CL) that penalizes the easy and hard cases of source examples. The CL approach is orthogonal to most of the UDA methods, including ours: it could benefit any method that uses cross-entropy for source.

Another approach for UDA is \textit{self-training}. The idea is to use the prediction from an ensembled model or a previous state of model as pseudo-labels for the unlabeled data to train the current model. Many semi-supervised methods \cite{laine2016temporal, tarvainen2017mean} use self-training. 
In \cite{zou2018unsupervised}, self-training is employed for UDA on semantic segmentation which is further extended with class balancing and spatial prior. Self-training has an interesting connection to the proposed entropy minimization approach as we discuss in Section \ref{sec:min_ent}.

Among some other approaches, \cite{murez2018CVPR} uses a combination of adversarial and generative techniques through multiple losses, \cite{zhang2018CVPR} combines the generative approach for appearance adaptation and adversarial training for representation adaptation, and  \cite{zhang2017curriculum} proposes a curriculum-style learning for UDA by enforcing the consistency on local (superpixel-level) and global label distributions.

\textit{Entropy minimization} has been shown to be useful for semi-supervised learning \cite{grandvalet2005semi,springenberg2015unsupervised} and clustering \cite{jain2017subic,jain2018learn}. It has also been recently applied on domain adaptation for classification task \cite{long2016unsupervised}. To our knowledge, we are first to successfully apply entropy based UDA training to obtain competitive performance on semantic segmentation task.

	\section{Approaches}\label{sec:approach}
	\begin{figure*}
	\centering
	\includegraphics[width=0.93\textwidth, trim={0.35cm 0.15cm 0 0cm},clip]{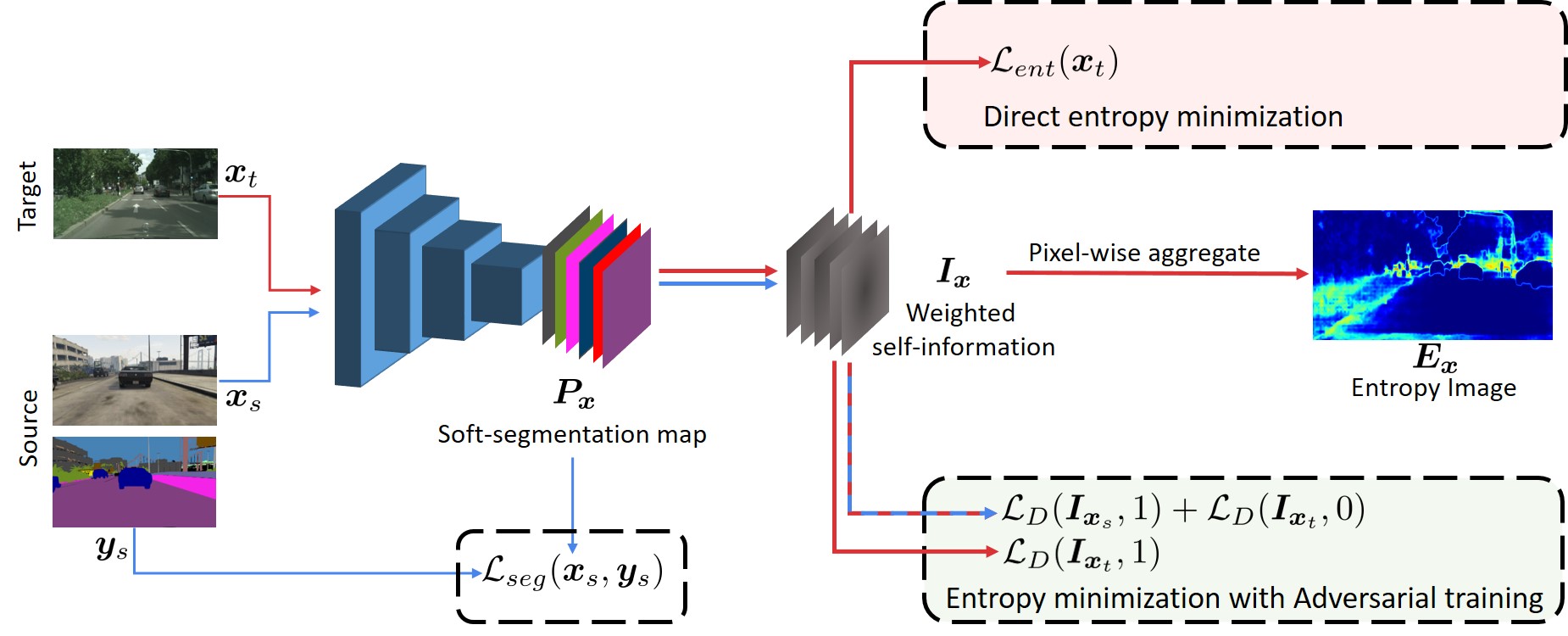}
	\vspace{-0.3cm}
	\caption{\small \textbf{Approach overview}. The figure shows our two approaches for UDA. First, \textit{direct entropy minimization} minimizes the entropy of the target $\mm{P}_{\trg{\mm{x}}}$, which is equivalent to minimizing the sum of weighted self-information maps $\mm{I}_{\trg{\mm{x}}}$. In the second, complementary approach, we use adversarial training to enforce the consistency in $\mm{I}_{\mm{x}}$ across domains. \textcolor{tcolor}{Red} arrows are used for target domain and \textcolor{scolor}{blue} arrows for source. An example of entropy map is shown for illustration.
	}
	\vspace{-0.5cm}
	\label{lbl:main_arch}
\end{figure*}

In this section, we present our two proposed approaches for entropy minimization using (i) an unsupervised entropy loss and (ii) adversarial training.
To build our models, we start from existing semantic segmentation frameworks and add an additional network branch used for domain adaptation.
Figure~\ref{lbl:main_arch} illustrates our architectures.

Our models are trained with a supervised loss on source domain. Formally, we consider a set $\src{\cX}\subset \mathbb{R}^{H\times W\times 3}$ of sources examples along with associated ground-truth $C$-class segmentation maps, $\src{\cY}\subset (1,C) ^{H\times W}$. Sample $\bx_s$ is a $H\times W$ color image and entry $\by_s^{(h,w)} = \big [\by_s^{(h,w,c)} \big]_c $ of associated map $\by_s$ provides the label of pixel $(h,w)$  as one-hot vector. 
Let $F$ be a semantic segmentation network which takes an image $\mm x$ and predicts a $C$-dimensional ``soft-segmentation map''
        $F(\bx) = \mm P_{\mm x} = \big[ \mm P_{\mm x}^{(h,w,c)} \big]_{h,w,c}$. By virtue of final softmax layer, each $C$-dimensional pixel-wise vector $\big[ \mm P_{\mm x}^{(h,w,c)} \big]_c$ behaves as a discrete distribution over classes. If one class stands out, the distribution is picky (low entropy), if scores are evenly spread, sign of uncertainty from the network standpoint, the entropy is large.
The parameters $\theta_F$ of $F$ are learned to minimize the segmentation loss $\mathcal{L}_{seg}(\src{\mm x}, \src{\mm y}) = -\sum_{h=1}^H\sum_{w=1}^W\sum_{c=1}^{C} \src{\mm y}^{(h,w,c)} \log \mm P_{\src{\mm x}}^{(h,w,c)}$
on source samples.
In the case of training only on source domain without domain adaptation, the optimization problem simply reads:
\begin{equation}\label{eq:onlysource}
\vspace{-0.1cm}
\min_{\theta_F} \frac{1}{|\cX_s|}\sum_{\mm x_s\in\cX_s} \mathcal{L}_{seg}(\src{\mm x}, \src{\mm y}).
\vspace{-0.1cm}
\end{equation}

\subsection{Direct entropy minimization}\label{sec:min_ent}
For the target domain, as we do not have the annotations $\trg{\mm{y}}$ for image samples $\bx_t\in\cX_t$, we cannot use \eqref{eq:onlysource} to learn $F$. Some methods use the model's prediction $\trg{\hat{\mm{y}}}$ as a proxy for $\trg{\mm{y}}$.
Also, this proxy is used only for pixels where prediction is sufficiently confident.
Instead of using the high-confident proxy, we propose to constrain the model such that it produces high-confident prediction. We realize this by minimizing the entropy of the prediction.

We introduce the entropy loss $\mathcal{L}_{ent}$ to directly maximize prediction certainty in the target domain.
In this work, we use the Shannon Entropy~\cite{shannon1948mathematical}.
Given a target input image $\trg{\mm x}$, the entropy map $\mm E_{\trg{\mm x}} \in [0,1]^{H\times W}$ is composed of the independent pixel-wise entropies normalized to $[0,1]$ range:
\begin{equation}\label{eq:eimage}
\vspace{-0.1cm}
\mm E_{\trg{\mm x}}^{(h,w)} = \frac{-1}{\log(C)}\sum_{c=1}^{C} \mm P_{\trg{\mm x}}^{(h,w,c)} \log \mm P_{\trg{\mm x}}^{(h,w,c)},
\vspace{-0.1cm}
\end{equation}
at pixel $(h,w)$. 
An example of entropy map is shown in Figure~\ref{lbl:main_arch}.
The entropy loss $\mathcal{L}_{ent}$ is defined as the sum of all pixel-wise normalized entropies:

\begin{equation}\label{eq:eloss}
\vspace{-0.1cm}
\mathcal{L}_{ent}(\trg{\mm x}) = \sum_{h,w} \mm E_{\trg{\mm x}}^{(h,w)}.
\vspace{-0.1cm}
\end{equation}
During training, we jointly optimize the supervised segmentation loss $\mathcal{L}_{seg}$ on source samples and the unsupervised entropy loss $\mathcal{L}_{ent}$ on target samples. 
The final optimization problem is formulated as follows:
\begin{equation}\label{eq:obj1}
\vspace{-0.1cm}
\min_{\theta_F} \frac{1}{|\cX_s|}\sum_{\bx_s} \mathcal{L}_{seg}(\src{\mm x}, \src{\mm y})  +  \frac{\lambda_{ent}}{|\cX_t|} \sum_{\bx_t}\mathcal{L}_{ent}(\trg{\mm x}),
\vspace{-0.1cm}
\end{equation}
with~$\lambda_{ent}$ as the weighting factor of the entropy term~$\mathcal{L}_{ent}$.

\vspace{-0.3cm}\paragraph{Connection to self-training.}
Pseudo-labeling is a simple yet efficient approach for semi-supervised learning~\cite{lee2013pseudo}.
Recently, the approach has been applied to UDA in semantic segmentation task with an iterative self-training (ST) procedure~\cite{zou2018unsupervised}.
The ST method assumes that the set $K \subset (1,H)\times (1,W)$ of high-scoring pixel-wise predictions on target samples are correct with high probability.
Such an assumption allows the use of cross-entropy loss with pseudo-labels on target predictions.
In practice, $K$ is constructed by selecting high-scoring pixels with a fixed or scheduled threshold.
To draw a link with entropy minimization, we write the training problem of the ST approach as:
\begin{equation}\label{eq:ST}
\vspace{-0.1cm}
\min_{\theta_F} \frac{1}{|\cX_s|}\sum_{\bx_s} \mathcal{L}_{seg}(\src{\mm x}, \src{\mm y})  + \frac{\lambda_{pl}}{|\cX_t|}\sum_{\bx_t}\mathcal{L}_{seg}(\trg{\mm x}, \trg{\hat{\by}}),
\vspace{-0.1cm}
\end{equation}
where $\hat{\by_t}$ is the one-hot class prediction for $\bx_t$ and with:
\begin{equation}\label{eq:ST_segloss}
\vspace{-0.1cm}
\mathcal{L}_{seg}(\trg{\mm x}, \trg{ \hat{\by}}) = -\sum_{(h,w)\in K}\sum_{c=1}^{C} \trg{\hat{\by}}^{(h,w,c)} \log\mm P_{\trg{\mm x}}^{(h,w,c)}.
\vspace{-0.1cm}
\end{equation}
Comparing equations~(\ref{eq:eimage}-\ref{eq:eloss}) and~\reff{eq:ST_segloss}, we note that our entropy loss $\mathcal{L}_{ent}(\trg{\mm x})$ can be seen as a soft-assignment version of the pseudo-label cross-entropy loss $\mathcal{L}_{seg}(\trg{\mm x}, \trg{\hat{\by}})$.
Different to ST~\cite{zou2018unsupervised}, our entropy-based approach does not require a complex scheduling procedure for choosing threshold.
Even, contrary to ST assumption, we show in Section~\ref{lbl:exp_abl} that, in some cases, training on the ``hard'' or ``most-confused'' pixels produces better performance.

\subsection{Minimizing entropy with adversarial learning} \label{sec:adv_ent}

The entropy loss for an input image is defined in equation~(\ref{eq:eloss}) as the sum of independent pixel-wise prediction entropies.
Therefore, a mere minimization of this loss neglects the structural dependencies between local semantics.
As shown in~\cite{tsai2018learning}, for UDA in semantic segmentation, adaptation on structured output space is beneficial.
It is based on the fact that source and target domain share strong similarities in semantic layout.

In this part, we introduce a unified adversarial training framework which indirectly minimizes the entropy by having target's entropy distribution similar to the source. This allows the exploitation of the structural consistency between the domains.  
To this end, we formulate the UDA task as minimizing distribution distance between source and target on the \textit{weighted self-information} space.
Figure~\ref{lbl:main_arch} illustrates our adversarial learning procedure.
Our adversarial approach is motivated by the fact that the trained model naturally produces low-entropy predictions on source-like images.
By aligning \THVT{weighted self-information distributions} of target and source domains, we indirectly minimize the entropy of target predictions.
Moreover, as the adaptation is done on the weighted self-information space, our model leverages structural information from source to target.

In detail, given a pixel-wise predicted class score $\mm P_{\mm x}^{(h,w,c)}$, the self-information or ``surprisal''~\cite{tribus1970thermostatics} is defined as $-\log \mm P_{\mm x}^{(h,w,c)}$.
Effectively, the entropy $\mm E_{\mm x}^{(h,w)}$ in~\reff{eq:eimage} is the expected value of the self-information $\E_{c}[-\log \mm P_{\mm x}^{(h,w,c)}]$.
We here perform adversarial adaptation on weighted self-information maps $\mm{I}_{\mm x}$ composed of pixel-level vectors $\mm{I}_{\mm x}^{(h,w)} = -\mm P_{\mm x}^{(h,w)}\cdot\log \mm P_{\mm x}^{(h,w)}.$\footnote{Abusing notations, 
'$\cdot$' and '$\log$' stand for Hadamard product and point-wise logarithm respectively.}
Such vectors can be seen as the disentanglement of the Shannon Entropy.
We then construct a fully-convolutional discriminator network $D$ with parameters $\theta_D$ taking $\mm I_{\mm x}$ as input and that produces domain classification outputs, \ie, class label $1$ (resp. $0$) for the source (resp. target) domain.
Similar to~\cite{goodfellow2014generative}, we train the discriminator to discriminate outputs coming from source and target images, and at the same time, train the segmentation network to fool the discriminator.
In detail, let $\mathcal{L}_{D}$ the cross-entropy domain classification loss.
The training objective of the discriminator is:
\begin{equation}\label{eq:objdis}
\min_{\theta_D} \frac{1}{|\cX_s|}\sum_{\bx_s}\mathcal{L}_{D}(\mm{I}_{\src{\mm x}}, 1) + \frac{1}{|\cX_t|}\sum_{\bx_t}\mathcal{L}_{D}(\mm{I}_{\trg{\mm x}}, 0),
\end{equation}
and the adversarial objective to train the segmentation network is:
\begin{equation}\label{eq:objadv}
\min_{\theta_F} \frac{1}{|\cX_t|}\sum_{\bx_t} \mathcal{L}_{D}(\mm{I}_{\trg{\mm x}}, 1).
\end{equation}
Combining~\reff{eq:onlysource} and~\reff{eq:objadv}, we derive the optimization problem
\begin{equation}\label{eq:objadvseg}
\min_{\theta_F} \frac{1}{|\cX_s|}\sum_{\bx_s}\mathcal{L}_{seg}(\src{\mm x}, \src{\mm y}) + \frac{\lambda_{adv}}{|\cX_t|}\sum_{\bx_t} \mathcal{L}_{D}(\mm{I}_{\trg{\mm x}}, 1),
\end{equation}
with the weighting factor $\lambda_{adv}$ for the adversarial term~$\mathcal{L}_{D}$.
During training, we alternatively optimize networks $D$ and $F$ using objective functions in  ~\reff{eq:objdis} and~\reff{eq:objadvseg}.

\subsection{Incorporating class-ratio priors}
\label{sec:class_prior}
Entropy minimization can get biased towards some easy classes. Therefore, sometimes it is beneficial to guide the learning with some prior. To this end, we use a simple class-prior based on the distribution of the classes over the source labels. We compute the class-prior vector $\mm p_s$ as a $\ell_1$-normalized histogram of number of pixels per class over the source labels. Now based on the predicted $\mm{P}_{\mm{x}_t}$, too large discrepancy between the expected probability for any class and class-prior $\mm p_s$ is penalized, using 
\begin{equation}
\mathcal{L}_{cp}(\mm{x}_t) = \sum_{c=1}^{C}\max\big(0, \mu\mm{p}_s^{(c)} - \E_c(\mm P_{\mm{x}_t}^{(c)})\big),
\end{equation}
where $\mu \in [0,1]$ is used to relax the class prior constraint.
This addresses the fact that class distribution on a single target image is not necessarily close to $\mm{p}_s$.

	\section{Experiments}
	\begin{table*}[t!]
	\scriptsize{
		\begin{center}
			\begin{tabular}{@{}L{2.5cm}@{}|@{}C{0.75cm}@{}|@{}C{0.7cm}@{}C{0.7cm}@{}C{0.7cm}@{}C{0.7cm}@{}C{0.7cm}@{}C{0.7cm}@{}C{0.7cm}@{}C{0.7cm}@{}C{0.7cm}@{}C{0.7cm}@{}C{0.7cm}@{}C{0.7cm}@{}C{0.7cm}@{}C{0.7cm}@{}C{0.7cm}@{}C{0.7cm}@{}C{0.7cm}@{}C{0.7cm}@{}C{0.7cm}@{}|@{}c@{}}
				\multicolumn{22}{c}{\small \rule{0pt}{2.5ex}(a) GTA5 $\rightarrow$ Cityscapes}\\
				\hline
				\hline
				Models & \rtb{Appr.} & \rtb{road} & \rtb{sidewalk\,\,} & \rtb{building} & \rtb{wall} & \rtb{fence} & \rtb{pole} & \rtb{light} & \rtb{sign} & \rtb{veg} & \rtb{terrain} & \rtb{sky} & \rtb{person} & \rtb{rider} & \rtb{car} & \rtb{truck} & \rtb{bus} & \rtb{train} & \rtb{mbike} & \rtb{bike} & \,mIoU\,\\
				\hline
				\rule{0pt}{3ex}FCNs in the Wild~\cite{hoffman2016fcns}&Adv&70.4&32.4&62.1&14.9&5.4&10.9&14.2&2.7&79.2&21.3&64.6&44.1&4.2&70.4&8.0&7.3&0.0&3.5&0.0&27.1\\
				\rule{0pt}{3ex}CyCADA~\cite{hoffman18a}&Adv&83.5&\textbf{38.3}&76.4&20.6&16.5&22.2&26.2&21.9&80.4&28.7&65.7&49.4&4.2&74.6&16.0&26.6&2.0&8.0&0.0&34.8\\
				\rule{0pt}{3ex}Adapt-SegMap~\cite{tsai2018learning}&Adv&87.3&29.8&78.6&21.1&18.2&22.5&21.5&11.0&79.7&29.6&71.3&46.8&6.5&80.1&23.0&26.9&0.0&10.6&0.3&35.0\\
				\rule{0pt}{3ex}Self-Training~\cite{zou2018unsupervised}&ST&83.8&17.4&72.1&14.3&2.9&16.5&16.0&6.8&81.4&24.2&47.2&40.7&7.6&71.7&10.2&7.6&0.5&11.1&0.9&28.1\\
				\rule{0pt}{3ex}Self-Training + CB~\cite{zou2018unsupervised}&ST&66.7&26.8&73.7&14.8&9.5&28.3&25.9&10.1&75.5&15.7&51.6&47.2&6.2&71.9&3.7&2.2&5.4&18.9&32.4&30.9\\
				\rowcolor[gray]{.92}[0pt][0pt]\rule{0pt}{3ex}Ours (MinEnt)&Ent&85.1&18.9&76.3&32.4&19.7&19.9&21.0&8.9&76.3&26.2&63.1&42.8&5.9&80.8&20.2&9.8&0.0&14.8&0.6&32.8\\
				\rowcolor[gray]{.92}[0pt][0pt]\rule{0pt}{3ex}Ours (AdvEnt)&Adv&86.9&28.7&78.7&28.5&25.2&17.1&20.3&10.9&80.0&26.4&70.2&47.1&8.4&81.5&26.0&17.2&\textbf{18.9}&11.7&1.6&36.1\\
				\hline
				\rule{0pt}{3ex}Adapt-SegMap~\cite{tsai2018learning}&Adv&86.5&36.0&79.9&23.4&23.3&23.9&35.2&14.8&83.4&33.3&75.6&58.5&27.6&73.7&32.5&35.4&3.9&{30.1}&28.1&42.4\\
				\rule{0pt}{3ex}Adapt-SegMap*&Adv&85.5&18.4&80.8&29.1&24.6&27.9&33.1&20.9&83.8&31.2&75.0&57.5&28.6&77.3&32.3&30.9&1.1&28.7&35.9&42.2\\
				\rowcolor[gray]{.92}[0pt][0pt]\rule{0pt}{3ex}Ours (MinEnt)&Ent&84.4&18.7&80.6&23.8&23.2&28.4&\textbf{36.9}&23.4&83.2&25.2&\textbf{79.4}&59.0&29.9&{78.5}&33.7&29.6&1.7&29.9&{33.6}&42.3\\
				\rowcolor[gray]{.92}[0pt][0pt]\rule{0pt}{3ex}Ours (MinEnt + ER)&Ent&84.2&25.2&77.0&17.0&23.3&24.2&33.3&\textbf{26.4}&80.7&32.1&{78.7}&57.5&30.0&77.0&37.9&44.3&1.8&31.4&\textbf{36.9}&{43.1}\\
				\rowcolor[gray]{.92}[0pt][0pt]\rule{0pt}{3ex}Ours (AdvEnt)&Adv&\textbf{89.9}&36.5&\textbf{81.6}&\textbf{29.2}&25.2&\textbf{28.5}&32.3&22.4&83.9&34.0&77.1&57.4&27.9&83.7&29.4&39.1&1.5&28.4&23.3&43.8\\
				\rowcolor[gray]{.92}[0pt][0pt]\rule{0pt}{3ex}Ours (AdvEnt+MinEnt)\,&A+E&89.4&33.1&{81.0}&{26.6}&\textbf{26.8}&27.2&{33.5}&24.7&\textbf{83.9}&\textbf{36.7}&{78.8}&\textbf{58.7}&\textbf{30.5}&\textbf{84.8}&\textbf{38.5}&\textbf{44.5}&1.7&\textbf{31.6}&32.4&\textbf{45.5}
			\end{tabular}
		\end{center}
		\vspace{-0.7cm}
		\begin{center}
			\begin{tabular}{@{}L{2.5cm}@{}|@{}C{0.8cm}@{}|@{}C{0.7cm}@{}C{0.7cm}@{}C{0.7cm}@{}C{0.7cm}@{}C{0.7cm}@{}C{0.7cm}@{}C{0.7cm}@{}C{0.7cm}@{}C{0.7cm}@{}C{0.7cm}@{}C{0.7cm}@{}C{0.7cm}@{}C{0.7cm}@{}C{0.7cm}@{}C{0.7cm}@{}C{0.7cm}@{}|@{}c@{}|@{}c@{}}
				\multicolumn{20}{c}{\small \rule{0pt}{2.5ex}(b) SYNTHIA $\rightarrow$ Cityscapes}\\
				\hline
				\hline
				Models & \rtb{Appr.} & \rtb{road} & \rtb{sidewalk\,\,} & \rtb{building} & \rtb{wall} & \rtb{fence} & \rtb{pole} & \rtb{light} & \rtb{sign} & \rtb{veg} & \rtb{sky} & \rtb{person} & \rtb{rider} & \rtb{car} & \rtb{bus} & \rtb{mbike} & \rtb{bike} & \,mIoU\, & \,mIoU*\,\\
				\hline
				\rule{0pt}{3ex}FCNs in the Wild~\cite{hoffman2016fcns}&Adv&11.5&19.6&30.8&4.4&0.0&20.3&0.1&11.7&42.3&68.7&51.2&3.8&54.0&3.2&0.2&0.6&20.2&22.1\\
				\rule{0pt}{3ex}Adapt-SegMap~\cite{tsai2018learning}&Adv&78.9&29.2&75.5&-&-&-&0.1&4.8&72.6&76.7&43.4&8.8&71.1&16.0&3.6&8.4&-&37.6\\
				\rule{0pt}{3ex}Self-Training~\cite{zou2018unsupervised}&ST&0.2&14.5&53.8&1.6&0.0&18.9&0.9&7.8&72.2&80.3&48.1&6.3&67.7&4.7&0.2&4.5&23.9&27.8\\
				\rule{0pt}{3ex}Self-Training + CB~\cite{zou2018unsupervised}&ST&69.6&28.7&69.5&\textbf{12.1}&0.1&25.4&\textbf{11.9}&\textbf{13.6}&\textbf{82.0}&81.9&49.1&14.5&66.0&6.6&3.7&32.4&35.4&36.1\\
				\rowcolor[gray]{.92}[0pt][0pt]\rule{0pt}{3ex}Ours (MinEnt)&Ent&37.8&18.2&65.8&2.0&0.0&15.5&0.0&0.0&76&73.9&45.7&11.3&66.6&13.3&1.5&13.1&27.5&32.5\\
				\rowcolor[gray]{.92}[0pt][0pt]\rule{0pt}{3ex}Ours (MinEnt + CP)&Ent&45.9&19.6&65.8&5.3&0.2&20.7&2.1&8.2&74.4&76.7&47.5&12.2&71.1&22.8&4.5&9.2&30.4&35.4\\
				\rowcolor[gray]{.92}[0pt][0pt]\rule{0pt}{3ex}Ours (AdvEnt + CP)&Adv&67.9&29.4&71.9&6.3&0.3&19.9&0.6&2.6&74.9&74.9&35.4&9.6&67.8&21.4&4.1&15.5&31.4&36.6\\
				\hline
				\rule{0pt}{3ex}Adapt-SegMap~\cite{tsai2018learning}&Adv&84.3&42.7&77.5&-&-&-&4.7&7.0&77.9&82.5&54.3&21.0&72.3&32.2&\textbf{18.9}&32.3&-&46.7\\
				\rule{0pt}{3ex}Adapt-SegMap*~\cite{tsai2018learning}&Adv&81.7&39.1&78.4&11.1&0.3&25.8&6.8&9.0&79.1&80.8&54.8&21.0&66.8&{34.7}&13.8&29.9&39.6&45.8\\
				\rowcolor[gray]{.92}[0pt][0pt]\rule{0pt}{3ex}Ours (MinEnt)&Ent&73.5&29.2&77.1&7.7&0.2&\textbf{27.0}&7.1&11.4&76.7&82.1&{57.2}&21.3&69.4&29.2&12.9&27.9&38.1&44.2\\
				\rowcolor[gray]{.92}[0pt][0pt]\rule{0pt}{3ex}Ours (AdvEnt)&Adv&\textbf{87.0}&\textbf{44.1}&\textbf{79.7}&9.6&\textbf{0.6}&24.3&4.8&7.2&80.1&{83.6}&56.4&{23.7}&{72.7}&32.6&12.8&\textbf{33.7}&{40.8}&{47.6}\\
				\rowcolor[gray]{.92}[0pt][0pt]\rule{0pt}{3ex}Ours (AdvEnt+MinEnt)&A+E&85.6&42.2&79.7&8.7&0.4&25.9&5.4&8.1&80.4&\textbf{84.1}&\textbf{57.9}&\textbf{23.8}&\textbf{73.3}&\textbf{36.4}&14.2&33.0&\textbf{41.2}&\textbf{48.0}
			\end{tabular}
		\end{center}
	}
	\vspace{-0.6cm}
	\caption{\small \textbf{Semantic segmentation performance mIoU (\%) on Cityscapes validation set of models trained on GTA5 (a) and SYNTHIA (b)}. We show results of our approaches using the direct entropy loss (MinEnt) and using adversarial training (AdvEnt). In each subtable, top and bottom parts correspond to VGG-16-based and ResNet-101-based models respectively. The ``Adapt-SegMap*'' denotes our retrained model of~\cite{tsai2018learning}. The abbreviations ``Adv'', ``ST'' and ``Ent'' stand for adversarial training, self-training and entropy minimization approaches.}
	\vspace{-0.5cm}
	\label{lbl:tbl_res}
\end{table*}

In this section, we present our experimental results.
Section~\ref{lbl:exp_detail} introduces the used datasets as well as our training parameters.
In Section~\ref{lbl:exp_result} and Section~\ref{lbl:exp_abl}, we report and discuss our main results.
\THVO{In Section~\ref{lbl:exp_abl}, we discuss a preliminary result on entropy-based UDA for detection.}

\subsection{Experimental details}
\label{lbl:exp_detail}
\vspace{-0.1cm}\paragraph{Datasets.}
To evaluate our approaches, we use the challenging \textit{synthetic-2-real} unsupervised domain adaptation set-ups.
Models are trained on fully-annotated synthetic data and are validated on real-world data.
In such set-ups, the models have access to some unlabeled real images during training.
To train our models, we use either GTA5~\cite{Richter_2016_ECCV} or SYNTHIA~\cite{Ros_2016_CVPR} as source domain synthetic data, along with the training split of Cityscapes dataset~\cite{cordts2016cityscapes} as target domain data.  Similar set-ups have been previously used in other works~\cite{hoffman2016fcns,hoffman18a,tsai2018learning,zou2018unsupervised}. In detail:
\vspace{-0.1cm}
\begin{itemize}
	\setlength{\parskip}{0pt}
	\setlength{\itemsep}{1pt}
	\item{\textit{GTA5$\rightarrow$Cityscapes:} The GTA5 dataset consists of $24,966$ synthesized frames captured from a video game. Images are provided with pixel-level semantic annotations of 33 classes. Similar to~\cite{hoffman2016fcns}, we use the $19$ classes in common with the Cityscapes dataset.}
	\item{\textit{SYNTHIA$\rightarrow$Cityscapes:} We use the SYNTHIA-RAND-CITYSCAPES set\footnote{\THVT{A split of the SYNTHIA dataset~\cite{Ros_2016_CVPR} using compatible labels with the Cityscapes dataset.}} with $9,400$ synthesized images for training. We train our models with $16$ common classes in SYNTHIA and Cityscapes. While evaluating we compare the performance on 16- and 13-class subsets following the protocol used in \cite{zou2018unsupervised}.}
\end{itemize}

In both set-ups, $2,975$ unlabeled Cityscapes images are used for training.
We measure segmentation performance with the standard mean-Intersection-over-Union (mIoU) metric~\cite{Everingham15}.
Evaluation is done on the $500$ validation images.

\vspace{-0.4cm}\paragraph{Network architectures.}
We use Deeplab-V2~\cite{chen2018deeplab} as the base semantic segmentation architecture $F$.
To better capture the scene context, Atrous Spatial Pyramid Pooling (ASPP) is applied on the last layer's feature output.
Sampling rates are fixed as $\{6, 12, 18, 24\}$, similar to the ASPP-L model in~\cite{chen2018deeplab}.
We experiment on the two different base deep CNN architectures: VGG-16~\cite{simonyan2014very} and ResNet-101~\cite{He2015}.
Following~\cite{chen2018deeplab}, we modify the stride and dilation rate of the last layers to produce denser feature maps with larger field-of-views.
To further improve performance on ResNet-101, we perform adaptation on multi-level outputs coming from both \textit{conv4} and \textit{conv5} features~\cite{tsai2018learning}.

The adversarial network $D$ introduced in Section~\ref{sec:adv_ent} has the same architecture as the one used in DCGAN~\cite{radford2015unsupervised}.
Weighted self-information maps $\mm{I}_{x}$ are forwarded through $4$ convolutional layers, each coupled with a leaky-ReLU layer with a fixed negative slope of $0.2$.
At the end, a classifier layer produces classification outputs, indicating if the inputs correspond to source or target domain.

\vspace{-0.4cm}\paragraph{Implementation details.}
We employ PyTorch deep learning framework~\cite{paszke2017automatic} in the implementations.
All experiments are done on a single NVIDIA 1080TI GPU with 11 GB memory.
Our model, except the adversarial discriminator mentioned in Section~\ref{sec:adv_ent}, is trained using Stochastic Gradient Descent optimizer~\cite{bottou2010large} with learning rate $2.5\times 10^{-4}$, momentum $0.9$ and weight decay $10^{-4}$.
We use Adam optimizer~\cite{kingma2014adam} with learning rate $10^{-4}$ to train the discriminator.
To schedule the learning rate, we follow the polynomial annealing procedure mentioned in~\cite{chen2018deeplab}.

\textit{Weighting factors} of entropy and adversarial losses: To set the weight for $\mathcal{L}_{ent}$, the training set performance provides important indications.
If $\lambda_{ent}$ is large then the entropy drops too quickly and the model is strongly biased towards a few classes.
When $\lambda_{ent}$ is chosen in a suitable range however, the performance is better and not sensitive to the precise value.
Thus, we use the same $\lambda_{ent}=0.001$ for all our experiments regardless of the network or the dataset.
Similar arguments hold for the weight~$\lambda_{adv}$ in~\reff{eq:objadvseg}.
We fix $\lambda_{adv}=0.001$ in all experiments.

\vspace{-0.05cm}\subsection{Results}
\label{lbl:exp_result}
We present experimental results of our approaches compared to different baselines. Our models achieve state-of-the-art performance in the two UDA benchmarks. In what follows, we show different behaviors of our approaches in different settings, \ie, training sets and base CNNs.

\vspace{-0.35cm}\paragraph{GTA5$\rightarrow$Cityscapes:} We report in Table~\ref{lbl:tbl_res}-a semantic segmentation performance in terms of mIoU (\%) on Cityscapes validation set.
Our first approach of direct entropy minimization, termed as MinEnt in Table~\ref{lbl:tbl_res}-a, achieves comparable performance to state-of-the-art baselines on both VGG-16 and ResNet-101-based CNNs.
The MinEnt outperforms Self-Training (ST) approach without and with class-balancing~\cite{zou2018unsupervised}.
Compared to~\cite{tsai2018learning}, the ResNet-101-based MinEnt shows similar results but without resorting to the training of a discriminator network.
The light overhead of the entropy loss makes training time much less for the MinEnt model.
Another advantage of our entropy approach is the ease of training.
Indeed, training adversarial networks is generally known as a difficult task due to its instability.
We observed a more stable behavior training models with the entropy loss. 

Interestingly, we find that in some cases, only applying entropy loss on certain ranges works best.
Such a phenomenon is observed with the ResNet-101-based models.
Indeed, we get a better model by training on pixels having entropy values within the top $30\%$ of each target sample.
The model is termed as MinEnt+ER in Table~\ref{lbl:tbl_res}-a.
We obtain $43.1\%$ mIoU using this strategy on the GTA5$\rightarrow$Cityscapes set-up.
More details are given in Section~\ref{lbl:exp_abl}.

Our second approach using adversarial training on the weighted self-information space, noted as AdvEnt, shows consistent improvement to the baselines on the two base networks.
In general, AdvEnt works better than MinEnt.
On the GTA5$\rightarrow$Cityscapes UDA set-up, AdvEnt achieves state-of-the-art mIoU of~$43.8$.
Such results confirm our intuition on the importance of structural adaptation.
With the VGG-16-based network, adaptation on the weighted self-information space brings~$+3.3\%$ mIoU improvement compared to the direct entropy minimization.
With the ResNet-101-based network, the improvement is less, \ie, $+1.5\%$ mIoU.
We conjecture that, as GTA5 semantic layouts are very similar to ones in Cityscapes, the segmentation network $F$ with high capacity base CNN like ResNet-101 is capable of learning some spatial priors from the supervision on source samples.
As for lower-capacity base model like VGG-16, an additional regularization on the structured space with adversarial training is more beneficial.

By combining results of the two models MinEnt and AdvEnt, we observe a decent boost in performance, compared to results of single models.
The ensemble achieves $45.5\%$ mIoU on the Cityscape validation set.
Such a result indicates that complementary information are learned by the two models.
\THVT{Indeed, while the entropy loss penalizes independent pixel-level predictions, the adversarial loss operates more on the image-level, \ie, scene topology.}
Similar to~\cite{tsai2018learning}, for a more meaningful comparison to other UDA approaches, in Table~\ref{tbl:perf_gap_oracle}-a we show the performance gap between UDA models and the oracle, \ie, the model trained with full supervision on the Cityscapes training set.
Compared to models trained by other methods, our single and ensemble models have smaller mIoU gaps to the oracle.

\begin{figure*}[t!]
	\vspace{-0.5cm}
	\begin{center}
		\begin{subfigure}[t]{0.24\textwidth}\centering
			\caption{Input image + GT}\vspace{-0.2cm}
			\includegraphics[width=.98\textwidth]{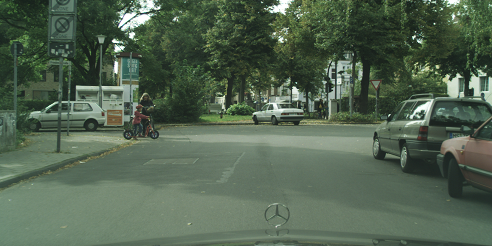}
		\end{subfigure}
		\begin{subfigure}[t]{0.24\textwidth}\centering
			\caption{Without adaptation}\vspace{-0.2cm}
			\includegraphics[width=.98\textwidth]{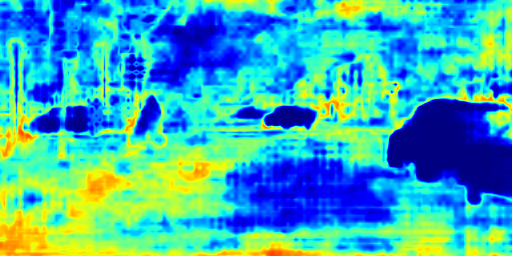}
		\end{subfigure}
		\begin{subfigure}[t]{0.24\textwidth}\centering
			\caption{MinEnt}\vspace{-0.2cm}
			\includegraphics[width=.98\textwidth]{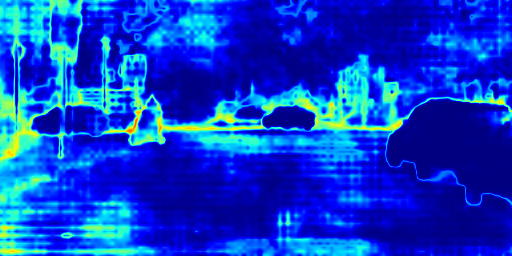}
		\end{subfigure}
		\begin{subfigure}[t]{0.24\textwidth}\centering
			\caption{AdvEnt}\vspace{-0.2cm}
			\includegraphics[width=.98\textwidth]{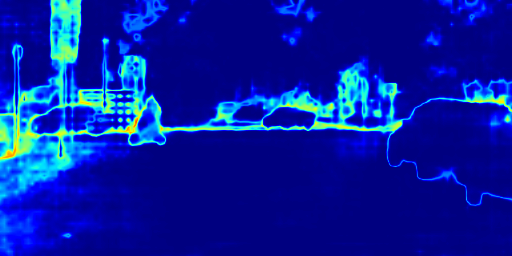}
		\end{subfigure}\\\vspace{0.1cm}
		\begin{subfigure}[t]{0.24\textwidth}\centering
			\includegraphics[width=.98\textwidth]{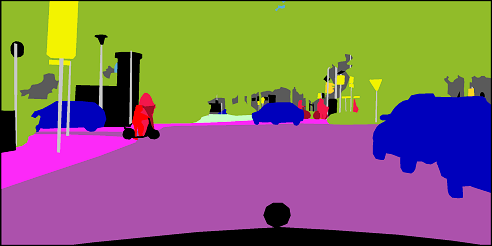}
		\end{subfigure}
		\begin{subfigure}[t]{0.24\textwidth}\centering
			\includegraphics[width=.98\textwidth]{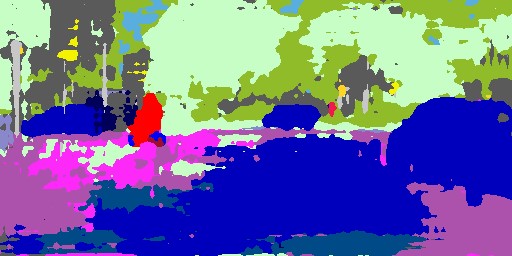}
		\end{subfigure}
		\begin{subfigure}[t]{0.24\textwidth}\centering
			\includegraphics[width=.98\textwidth]{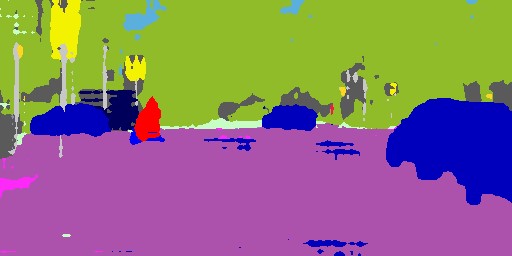}
		\end{subfigure}
		\begin{subfigure}[t]{0.24\textwidth}\centering
			\includegraphics[width=.98\textwidth]{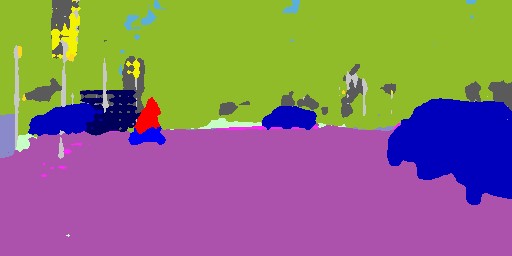}
		\end{subfigure}\\\vspace{-0.13cm}
		\hdashrule[1ex][x]{17cm}{1.5pt}{1.5mm}\vspace{-0.13cm}
		\begin{subfigure}[t]{0.24\textwidth}\centering
			\includegraphics[width=.98\textwidth]{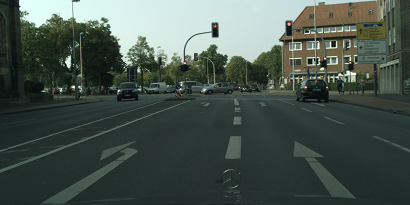}
		\end{subfigure}
		\begin{subfigure}[t]{0.24\textwidth}\centering
			\includegraphics[width=.98\textwidth]{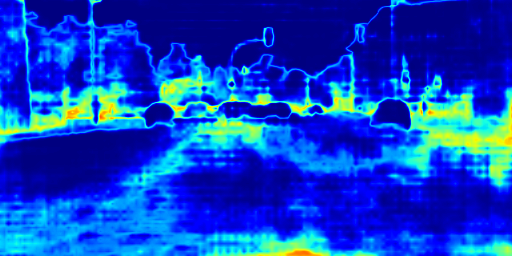}
		\end{subfigure}
		\begin{subfigure}[t]{0.24\textwidth}\centering
			\includegraphics[width=.98\textwidth]{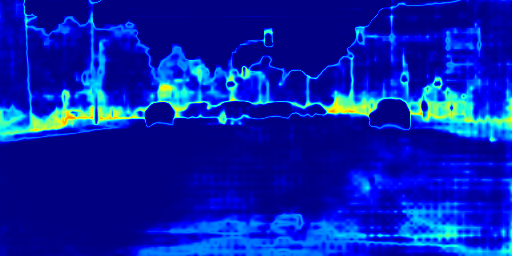}
		\end{subfigure}
		\begin{subfigure}[t]{0.24\textwidth}\centering
			\includegraphics[width=.98\textwidth]{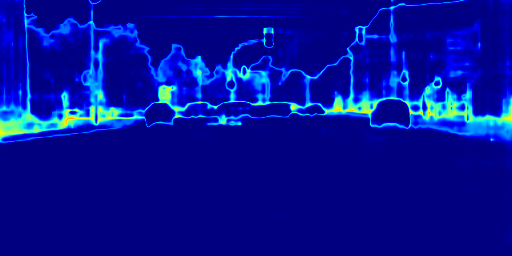}
		\end{subfigure}\\\vspace{0.1cm}
		\begin{subfigure}[t]{0.24\textwidth}\centering
			\includegraphics[width=.98\textwidth]{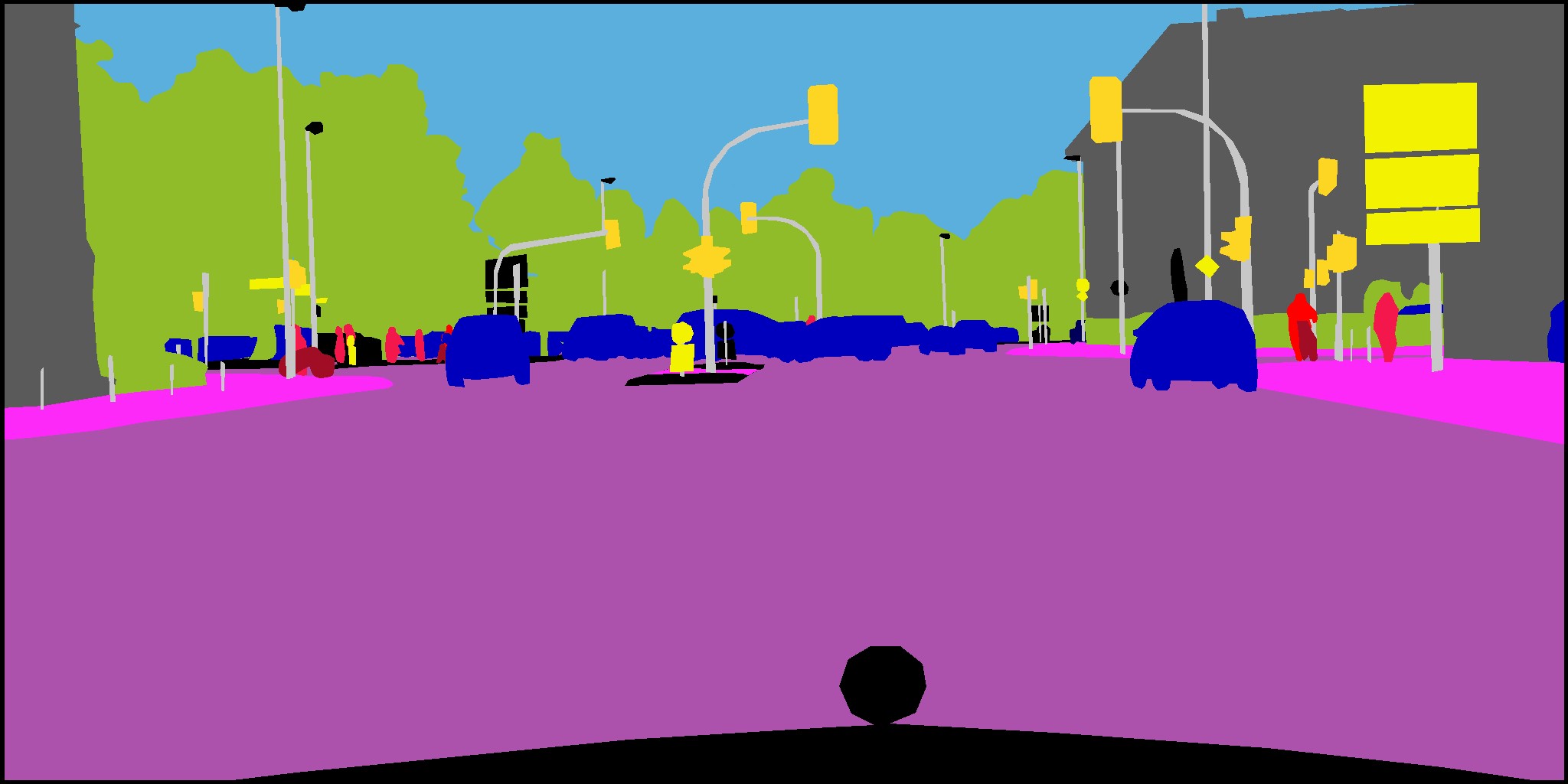}
		\end{subfigure}
		\begin{subfigure}[t]{0.24\textwidth}\centering
			\includegraphics[width=.98\textwidth]{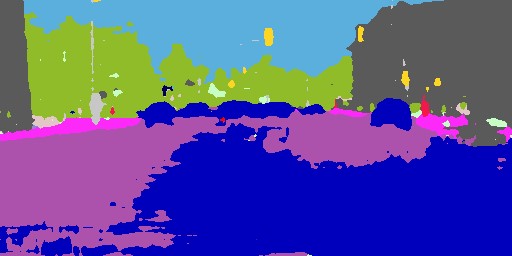}
		\end{subfigure}
		\begin{subfigure}[t]{0.24\textwidth}\centering
			\includegraphics[width=.98\textwidth]{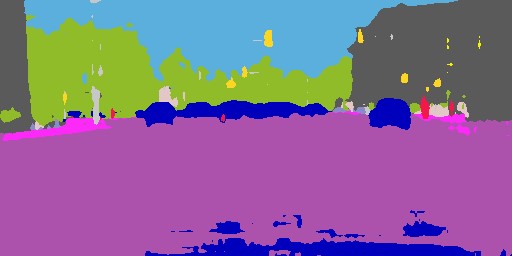}
		\end{subfigure}
		\begin{subfigure}[t]{0.24\textwidth}\centering
			\includegraphics[width=.98\textwidth]{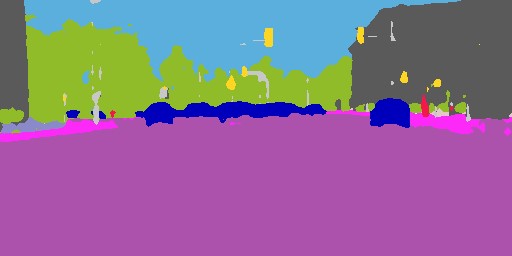}
		\end{subfigure}\\\vspace{-0.13cm}
		\hdashrule[1ex][x]{17cm}{1.5pt}{1.5mm}\vspace{-0.13cm}
		\begin{subfigure}[t]{0.24\textwidth}\centering
			\includegraphics[width=.98\textwidth]{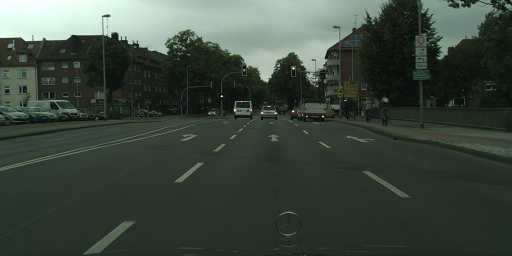}
		\end{subfigure}
		\begin{subfigure}[t]{0.24\textwidth}\centering
			\includegraphics[width=.98\textwidth]{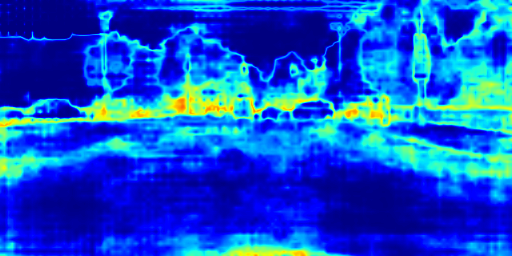}
		\end{subfigure}
		\begin{subfigure}[t]{0.24\textwidth}\centering
			\includegraphics[width=.98\textwidth]{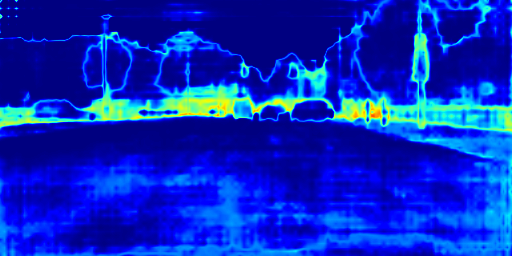}
		\end{subfigure}
		\begin{subfigure}[t]{0.24\textwidth}\centering
			\includegraphics[width=.98\textwidth]{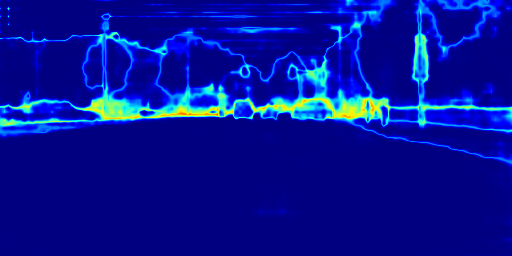}
		\end{subfigure}\\\vspace{0.1cm}
		\begin{subfigure}[t]{0.24\textwidth}\centering
			\includegraphics[width=.98\textwidth]{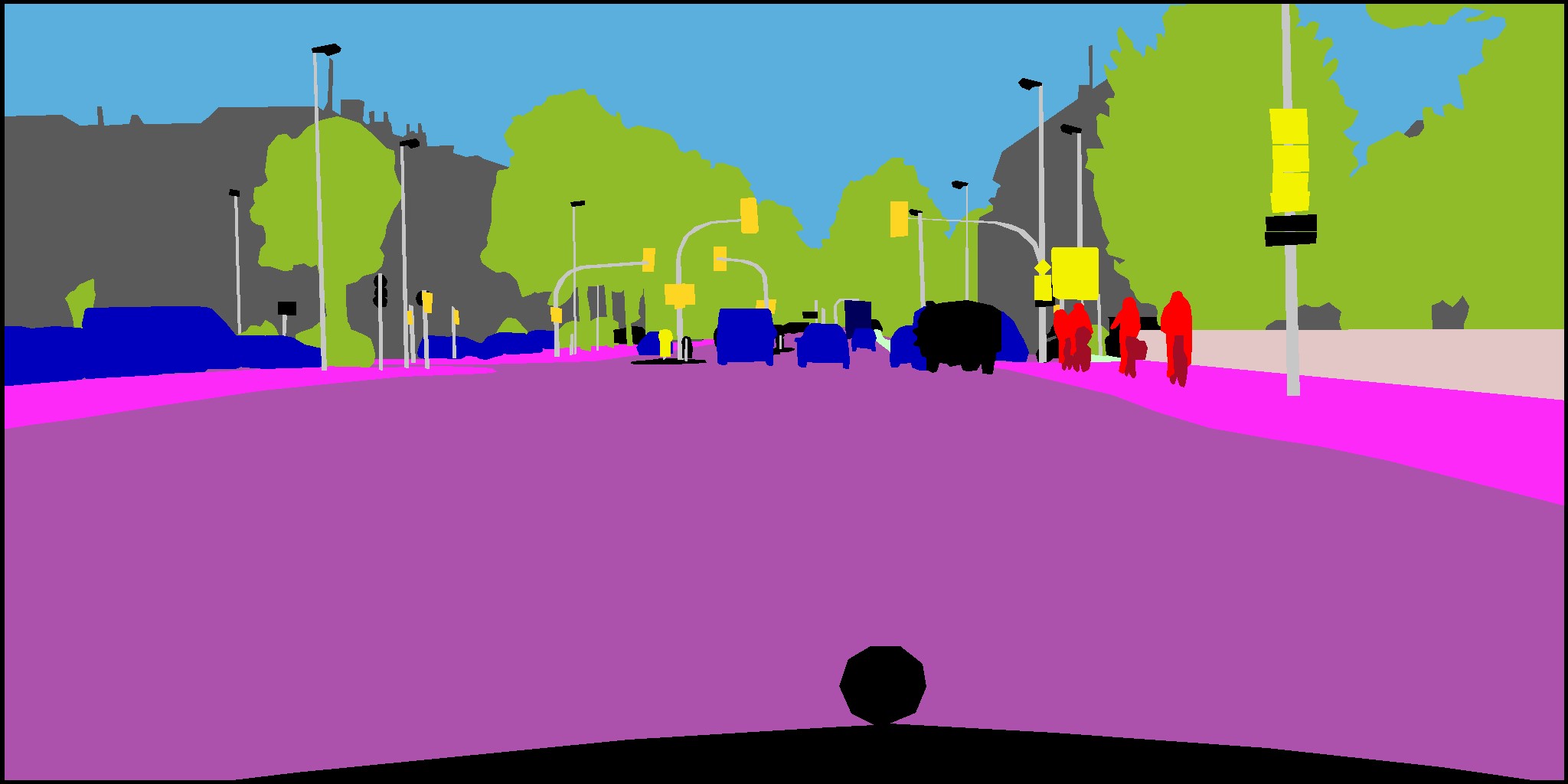}
		\end{subfigure}
		\begin{subfigure}[t]{0.24\textwidth}\centering
			\includegraphics[width=.98\textwidth]{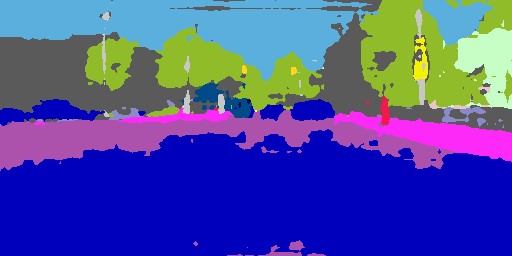}
		\end{subfigure}
		\begin{subfigure}[t]{0.24\textwidth}\centering
			\includegraphics[width=.98\textwidth]{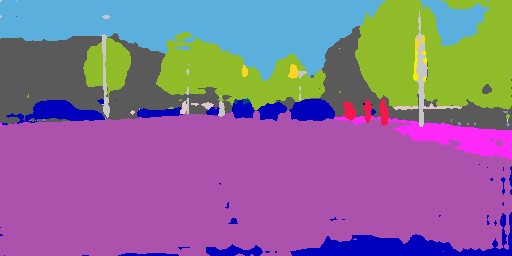}
		\end{subfigure}
		\begin{subfigure}[t]{0.24\textwidth}\centering
			\includegraphics[width=.98\textwidth]{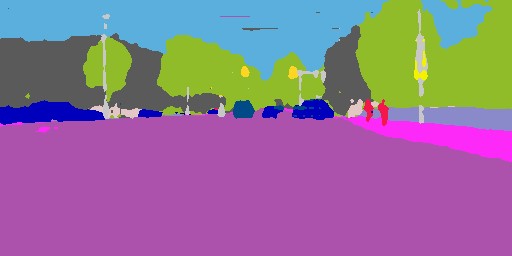}
		\end{subfigure}\\\vspace{-0.13cm}
		\hdashrule[1ex][x]{17cm}{1.5pt}{1.5mm}\vspace{-0.13cm}
		\begin{subfigure}[t]{0.24\textwidth}\centering
			\includegraphics[width=.98\textwidth]{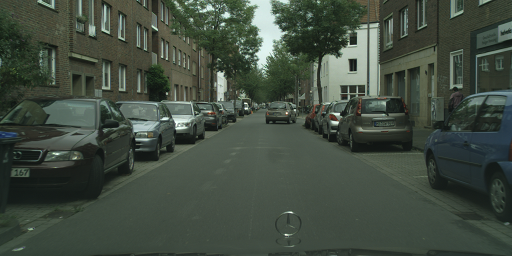}
		\end{subfigure}
		\begin{subfigure}[t]{0.24\textwidth}\centering
			\includegraphics[width=.98\textwidth]{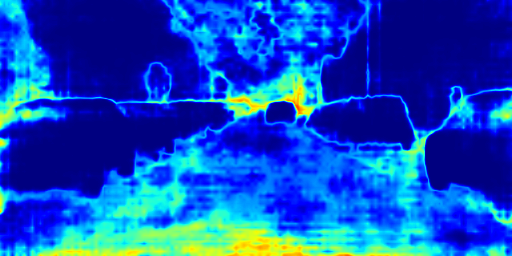}
		\end{subfigure}
		\begin{subfigure}[t]{0.24\textwidth}\centering
			\includegraphics[width=.98\textwidth]{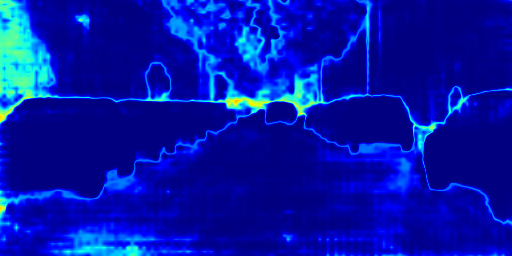}
		\end{subfigure}
		\begin{subfigure}[t]{0.24\textwidth}\centering
			\includegraphics[width=.98\textwidth]{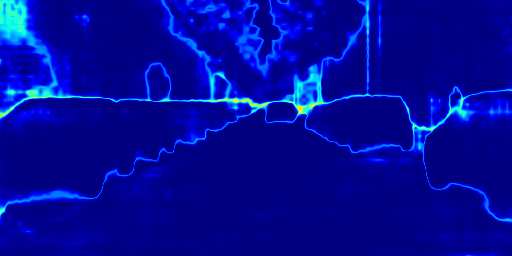}
		\end{subfigure}\\\vspace{0.1cm}
		\begin{subfigure}[t]{0.24\textwidth}\centering
			\includegraphics[width=.98\textwidth]{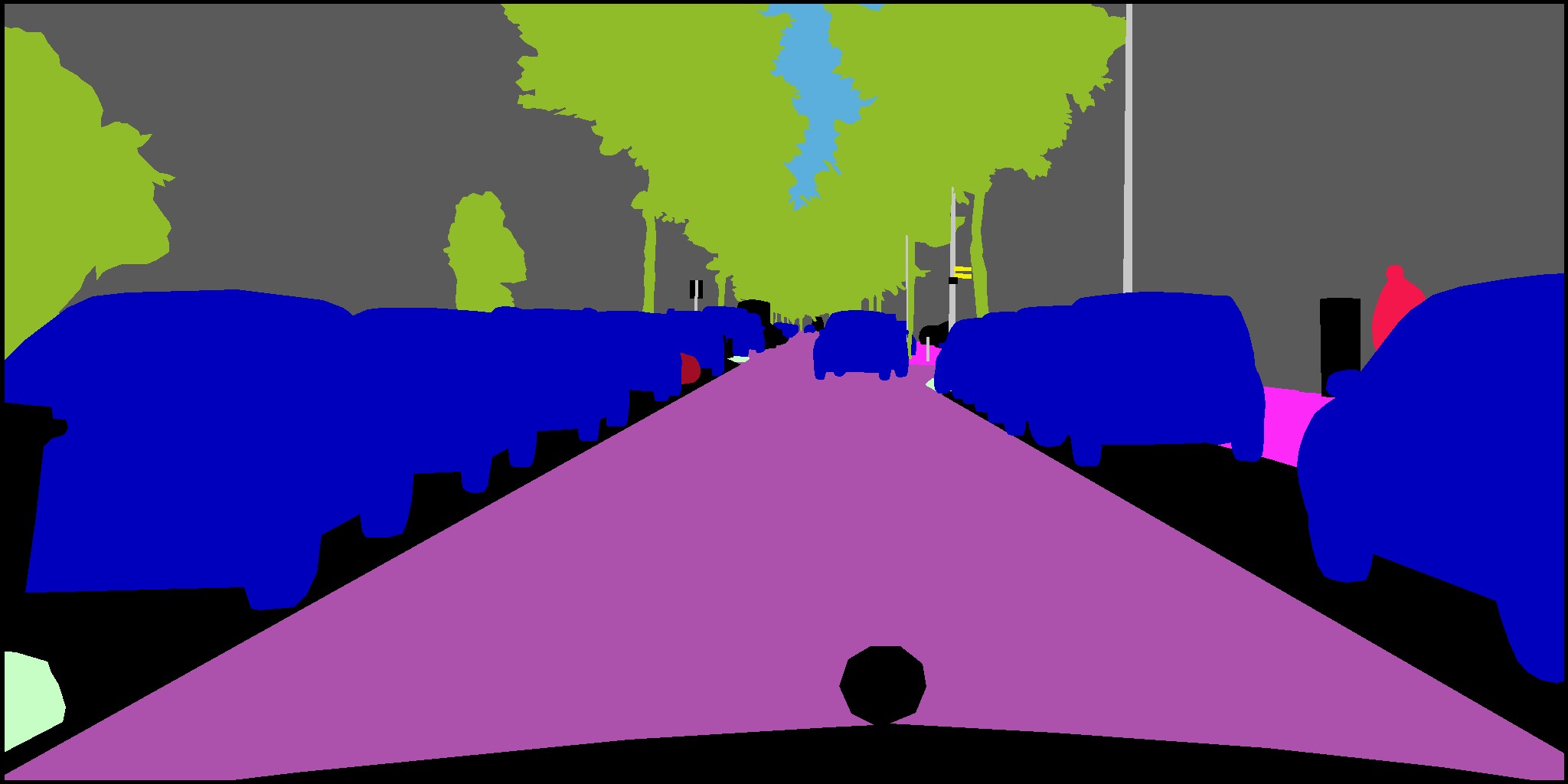}
		\end{subfigure}
		\begin{subfigure}[t]{0.24\textwidth}\centering
			\includegraphics[width=.98\textwidth]{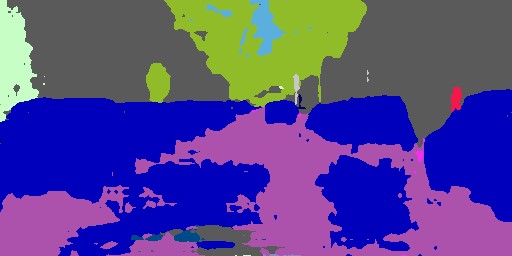}
		\end{subfigure}
		\begin{subfigure}[t]{0.24\textwidth}\centering
			\includegraphics[width=.98\textwidth]{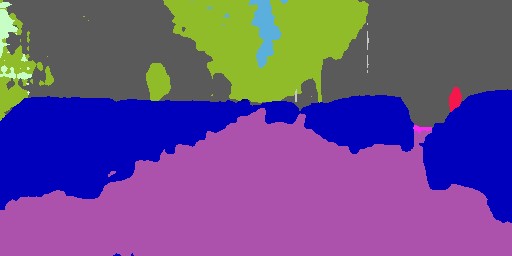}
		\end{subfigure}
		\begin{subfigure}[t]{0.24\textwidth}\centering
			\includegraphics[width=.98\textwidth]{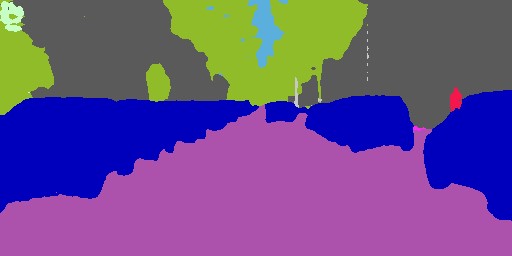}
		\end{subfigure}\\\vspace{-0.13cm}
		\hdashrule[1ex][x]{17cm}{1.5pt}{1.5mm}\vspace{-0.13cm}
		\begin{subfigure}[t]{0.24\textwidth}\centering
			\includegraphics[width=.98\textwidth]{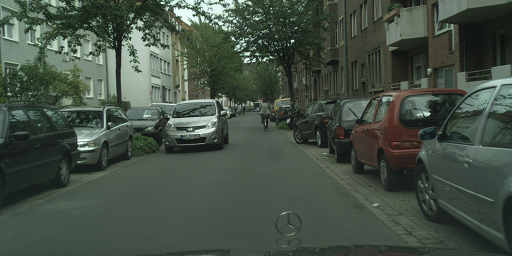}
		\end{subfigure}
		\begin{subfigure}[t]{0.24\textwidth}\centering
			\includegraphics[width=.98\textwidth]{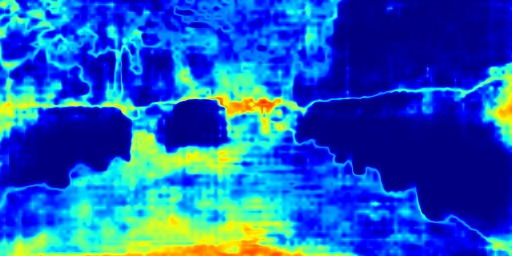}
		\end{subfigure}
		\begin{subfigure}[t]{0.24\textwidth}\centering
			\includegraphics[width=.98\textwidth]{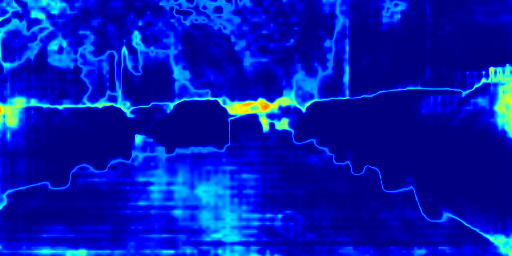}
		\end{subfigure}
		\begin{subfigure}[t]{0.24\textwidth}\centering
			\includegraphics[width=.98\textwidth]{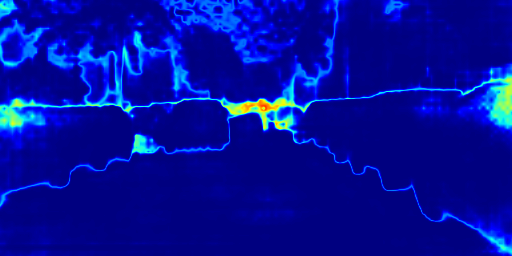}
		\end{subfigure}\\\vspace{0.1cm}
		\begin{subfigure}[t]{0.24\textwidth}\centering
			\includegraphics[width=.98\textwidth]{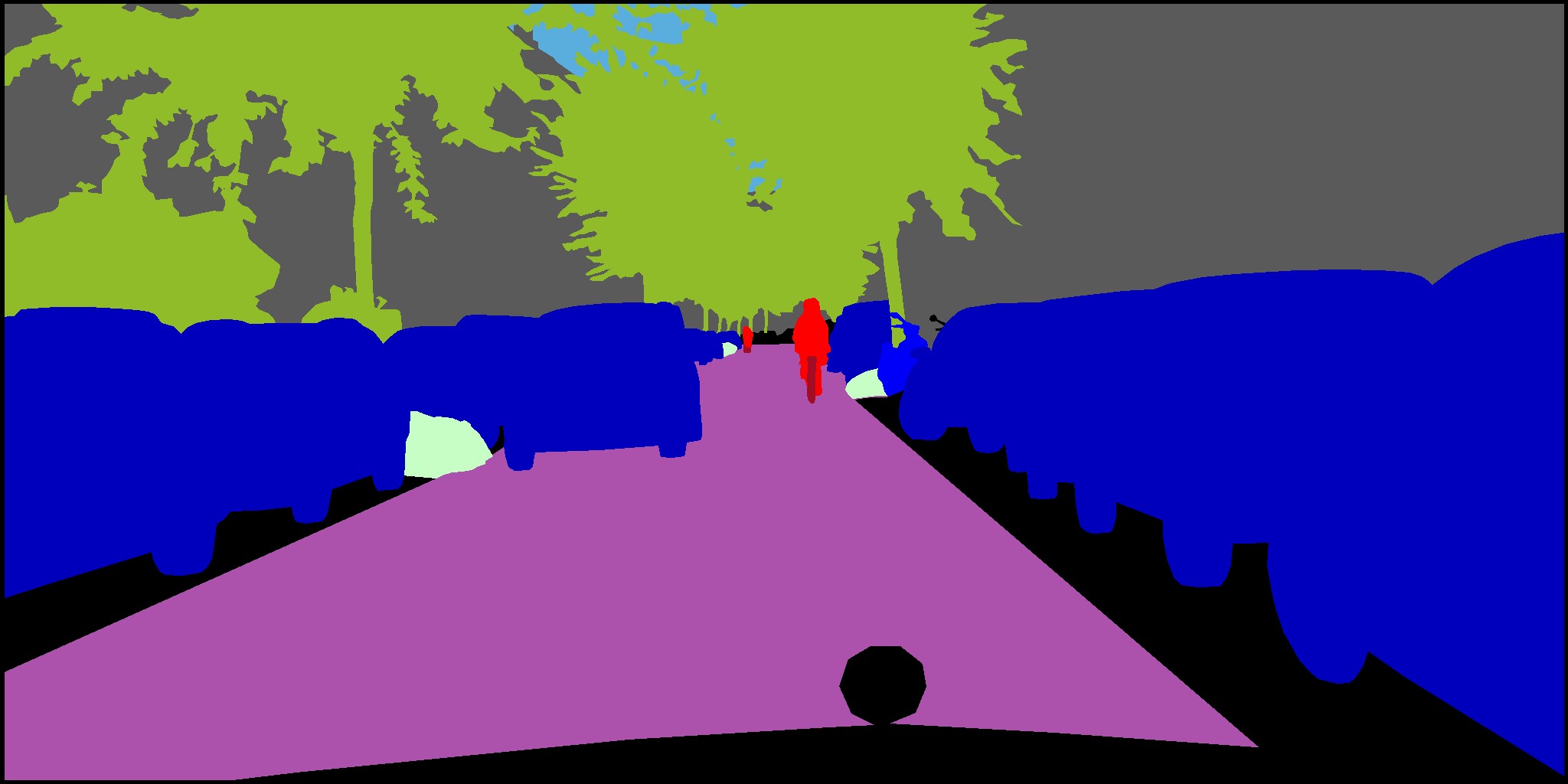}
		\end{subfigure}
		\begin{subfigure}[t]{0.24\textwidth}\centering
			\includegraphics[width=.98\textwidth]{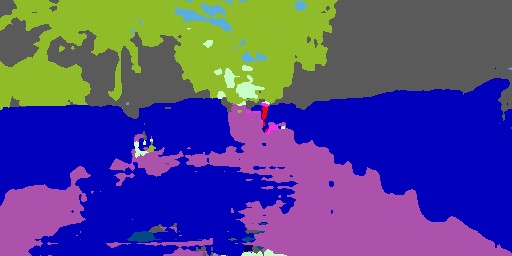}
		\end{subfigure}
		\begin{subfigure}[t]{0.24\textwidth}\centering
			\includegraphics[width=.98\textwidth]{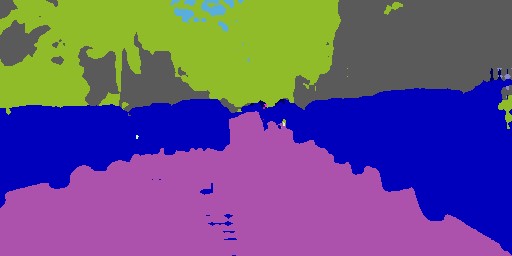}
		\end{subfigure}
		\begin{subfigure}[t]{0.24\textwidth}\centering
			\includegraphics[width=.98\textwidth]{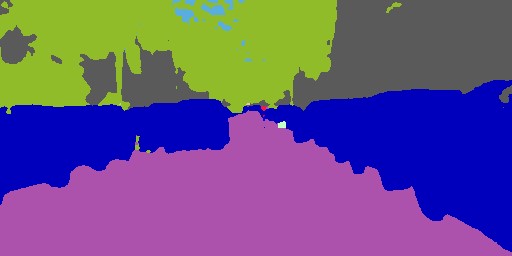}
		\end{subfigure}
	\end{center}
	\vspace{-0.3cm}
	\caption{\small \textbf{Qualitative results in the GTA5$\rightarrow$Cityscapes set-up}. Column (a) shows a input image and the corresponding semantic segmentation ground-truth. Column (b), (c) and (d) show segmentation results (bottom) along with prediction entropy maps produced by different approaches (top). Best viewed in colors.}
	\vspace{-0.4cm}
	\label{fig:sup_qual_seg}
\end{figure*}

\THVT{In Figure~\ref{fig:sup_qual_seg}, we illustrate some qualitative results of our models.
	Without domain adaptation, the model trained only on source supervision produces noisy segmentation predictions as well as high entropy activations, with a few exceptions on some classes like ``building'' and ``car''.
	Still, there exist many confident predictions (low entropy) which are completely wrong.
	Our models, on the other hand, manage to produce correct predictions at high level of confidence.
	We observe that overall, the AdvEnt model achieves lower prediction entropy compared to the MinEnt model.
}

\begin{table}[t!]
	\begin{center}
		\setlength{\tabcolsep}{5pt}
		\renewcommand{\arraystretch}{.6}
		\scriptsize{
			\begin{tabular}{l|cccc}
				\multicolumn{4}{c}{\small \rule{0pt}{2.5ex} (a) GTA5 $\rightarrow$ Cityscapes}\\
				\hline
				\hline
				\rule{0pt}{3ex}Method&UDA Model&Oracle&mIoU Gap (\%)\\
				\hline
				\rule{0pt}{3ex}FCNs in the Wild~\cite{hoffman2016fcns} & 27.1 & 64.6 & -37.5 \\
				\rule{0pt}{3ex}CyCADA~\cite{hoffman18a} & 28.9 & 60.3 & -31.4 \\
				\rule{0pt}{3ex}Adapt-SegNet~\cite{tsai2018learning} & 35.0 & 61.8 & -26.8 \\
				\rowcolor[gray]{.92}\rule{0pt}{3ex}Ours (single model) & 36.1 & 61.8 & -25.7\\
				\hline
				\rule{0pt}{3ex}Adapt-SegNet~\cite{tsai2018learning} & 42.4 & 65.1 & -22.7 \\
				\rowcolor[gray]{.92}\rule{0pt}{3ex}Ours (single model) & 43.8 & 65.1 & {-21.3}\\
				\rowcolor[gray]{.92}\rule{0pt}{3ex}Ours (two models) & 45.5 & 65.1 & \textbf{-19.6}\\
			\end{tabular}
		}
	\end{center}
	\vspace{-0.7cm}
	\begin{center}
		\setlength{\tabcolsep}{5pt}
		\renewcommand{\arraystretch}{.6}
		\scriptsize{
			\begin{tabular}{l|cccc}
				\multicolumn{4}{c}{\small \rule{0pt}{2.5ex} (b) SYNTHIA $\rightarrow$ Cityscapes}\\
				\hline
				\hline
				\rule{0pt}{3ex}Method&UDA Model&Oracle&mIoU Gap (\%)\\
				\hline
				\rule{0pt}{3ex}FCNs in the Wild~\cite{hoffman2016fcns} & 22.9 & 73.8 & -50.9 \\
				\rule{0pt}{3ex}Adapt-SegNet~\cite{tsai2018learning} & 37.6 & 68.4 & -30.8 \\
				\rowcolor[gray]{.92}\rule{0pt}{3ex}Ours (single model) & 36.6 & 68.4 & -31.8\\
				\hline
				\rule{0pt}{3ex}Adapt-SegNet~\cite{tsai2018learning} & 46.7 & 71.7 & -25.0 \\
				\rowcolor[gray]{.92}\rule{0pt}{3ex}Ours (single model) & 47.6 & 71.7 & {-24.1}\\
				\rowcolor[gray]{.92}\rule{0pt}{3ex}Ours (two models) & 48 & 71.7 & \textbf{-23.7}\\
			\end{tabular}
		}
	\end{center}
	\vspace{-0.5cm}
	\caption{\small\textbf{ Performance gap between UDA models and the oracle} in GTA5$\rightarrow$Cityscapes and SYNTHIA$\rightarrow$Cityscapes setups. Top and bottom parts of each table correspond to VGG-16-based and ResNet-101-based models respectively.}
	\vspace{-0.5cm}
	\label{tbl:perf_gap_oracle}
\end{table}

\vspace{-0.35cm}\paragraph{SYNTHIA$\rightarrow$Cityscapes:} Table~\ref{lbl:tbl_res}-b shows results on the~$16$- and~$13$-class subsets of the Cityscapes validation set.
We notice that scene images in SYNTHIA cover more diverse viewpoints than the ones in GTA5 and Cityscape.
This results in different behaviors of our approaches.

On the VGG-16-based network, the MinEnt model shows comparable results to state-of-the-art methods.
Compared to Self-Training~\cite{zou2018unsupervised}, our model achieves~$+3.6\%$ and~$+4.7\%$ on $16$- and $13$- class subsets respectively.
However, compared to stronger baselines like the class-balanced self-training, we observe a significant drop in class ``road''.
We argue that it is due to the large layout gaps between SYNTHIA and Cityscapes.
To target this issue, we incorporate the class-ratio priors from source domain, as introduced in Section~\ref{sec:class_prior}.
By constraining target output distribution using class-ratio priors, noted as CP in Table~\ref{lbl:tbl_res}-b, we improve MinEnt by~$+2.9\%$ mIoU on both $16$- and $13$- class subsets.
With adversarial training, we have an additional~$\sim+1\%$ mIoU.
On the ResNet-101-based network, the AdvEnt model achieves state-of-the-art performance.
Compared to the retrained model of~\cite{tsai2018learning}, \ie, Adapt-SegMap*, the AdvEnt improves the mIoUs on $16$- and $13$- class subsets by~$+1.2\%$ and~$+1.8\%$.

Consistent with the GTA5 results above, the ensemble of the two models MinEnt and AdvEnt trained on SYNTHIA achieves the best mIoU of $41.2\%$ and $48.0\%$ on $16$- and $13$- class subsets respectively.
According to Table~\ref{tbl:perf_gap_oracle}-b, our models have the smallest mIoU gaps to the oracle.

\subsection{Discussion}
\label{lbl:exp_abl}
The experimental results shown in Section~\ref{lbl:exp_result} have validated the advantages of our approaches.
To further push the performance, we proposed two different ways to regularize the training in two particular settings.
This section discusses our experimental choices and explain the intuitions behind.

\vspace{-0.4cm}\paragraph{GTA5$\rightarrow$Cityscapes: Training on specific entropy ranges.}
In this setup, we observe that the performance of model MinEnt using ResNet-101-based network can be improved by training on target pixels having entropy values in a specific range.
Interestingly, the best MinEnt model was trained on the top $30\%$ highest-entropy pixels on each target sample -- gaining $0.8\%$ mIoU over the vanilla model.
We note that high-entropy pixels are the ``most confusing'' ones, \ie, where the segmentation model is indecisive between multiple classes.
One reason is that the ResNet-101-based model generalizes well in this particular setting.
Accordingly, among the ``most confusing'' predictions, there is a decent amount of correct but ``low confident'' ones.
Minimizing the entropy loss on such a set still drives the model toward the desirable direction.
This assumption, however, does not hold for the VGG-16-based model.

\vspace{-0.45cm}\paragraph{SYNTHIA$\rightarrow$Cityscapes: Using class-ratio prior.}
As discussed before, SYNTHIA has significantly different layout and viewpoints than Cityscapes. This disparity can cause very bad prediction for some classes, which is then further encouraged with minimization of entropy or their use as label proxy in self-training. Thus, it can lead to strong class biases or, in extreme cases, to missing some classes completely in the target domain.
Adding class-ratio prior encourages the presence of all the classes and thus helps avoid such degenerate solutions. As mentioned in Sec. \ref{sec:class_prior}, we use $\mu$ to relax the source class-ratio prior, for example $\mu=0$ means no prior while $\mu=1$ implies enforcing exact class-ratio prior. Having $\mu=1$ is not ideal as it means that each target image should follow the class-ratio from the source domain. We choose $\mu=0.5$ to let the target class-ratio to be somewhat different from the source.

\begin{figure*}[t!]
	\vspace{-0.5cm}
	\begin{center}
		\begin{subfigure}[t]{0.24\textwidth}\centering
			\caption{Input image}\vspace{-0.2cm}
			\includegraphics[width=.98\textwidth]{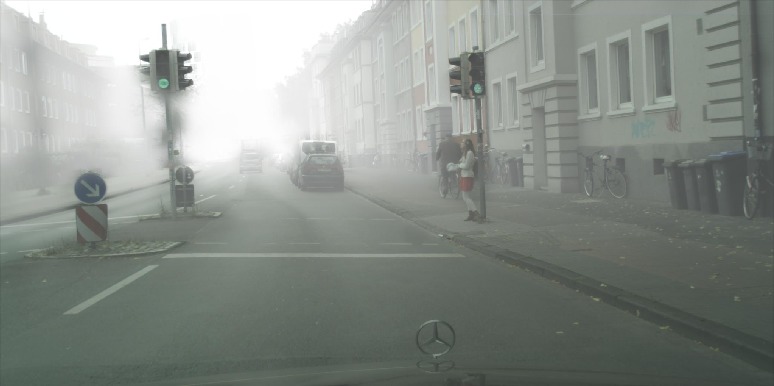}
		\end{subfigure}
		\begin{subfigure}[t]{0.24\textwidth}\centering
			\caption{Without adaptation}\vspace{-0.2cm}
			\includegraphics[width=.98\textwidth]{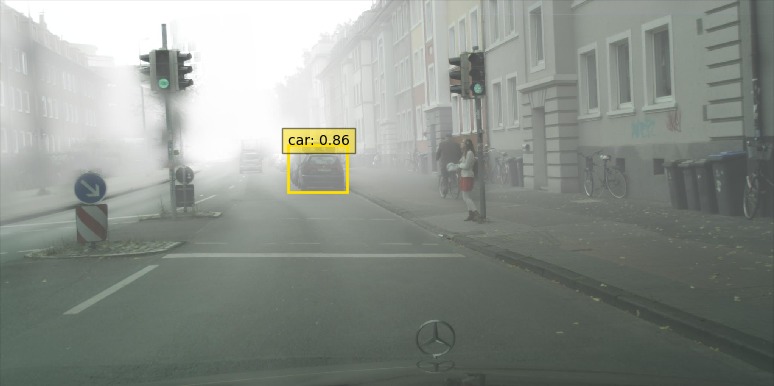}
		\end{subfigure}
		\begin{subfigure}[t]{0.24\textwidth}\centering
			\caption{MinEnt}\vspace{-0.2cm}
			\includegraphics[width=.98\textwidth]{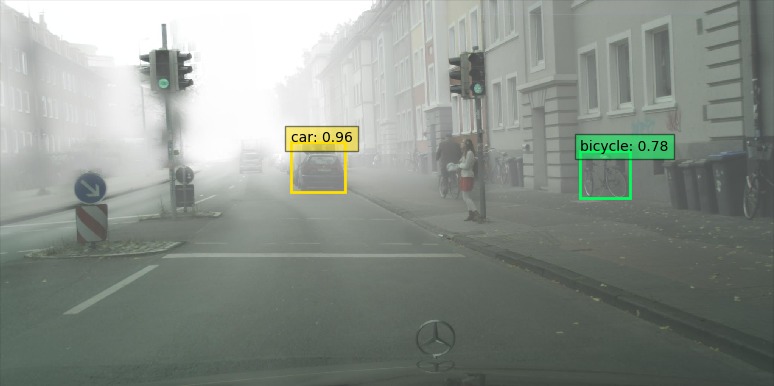}
		\end{subfigure}
		\begin{subfigure}[t]{0.24\textwidth}\centering
			\caption{AdvEnt}\vspace{-0.2cm}
			\includegraphics[width=.98\textwidth]{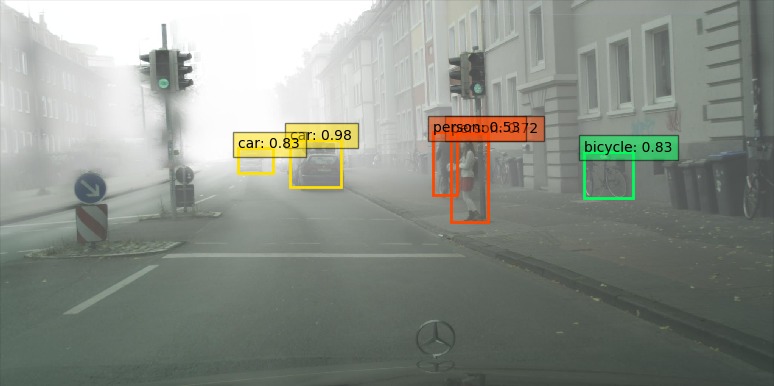}
		\end{subfigure}\\\vspace{0.1cm}
		\begin{subfigure}[t]{0.24\textwidth}\centering
			\includegraphics[width=.98\textwidth]{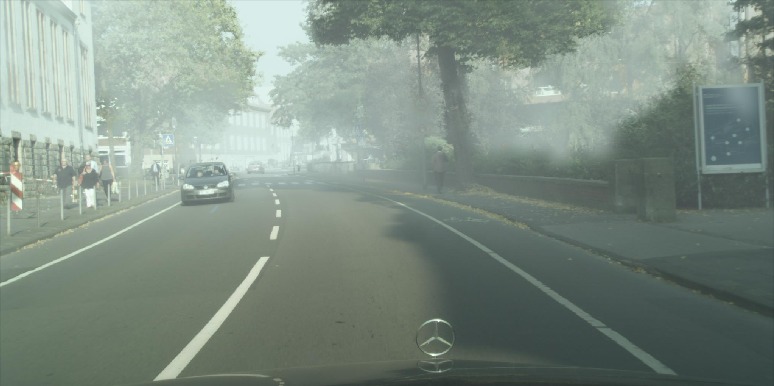}
		\end{subfigure}
		\begin{subfigure}[t]{0.24\textwidth}\centering
			\includegraphics[width=.98\textwidth]{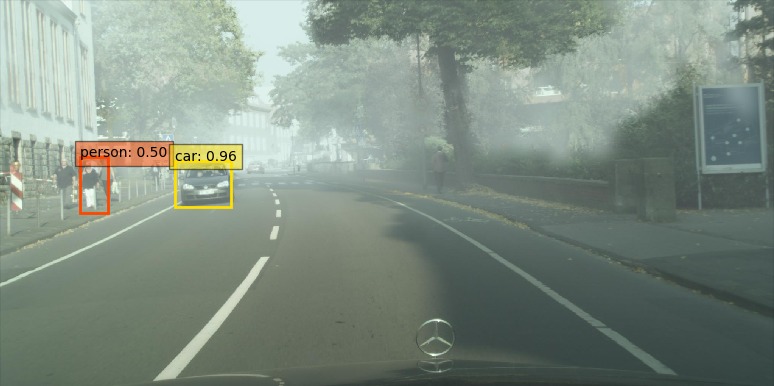}
		\end{subfigure}
		\begin{subfigure}[t]{0.24\textwidth}\centering
			\includegraphics[width=.98\textwidth]{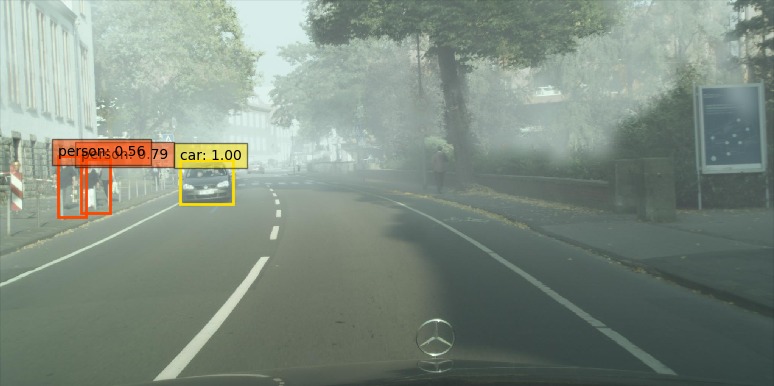}
		\end{subfigure}
		\begin{subfigure}[t]{0.24\textwidth}\centering
			\includegraphics[width=.98\textwidth]{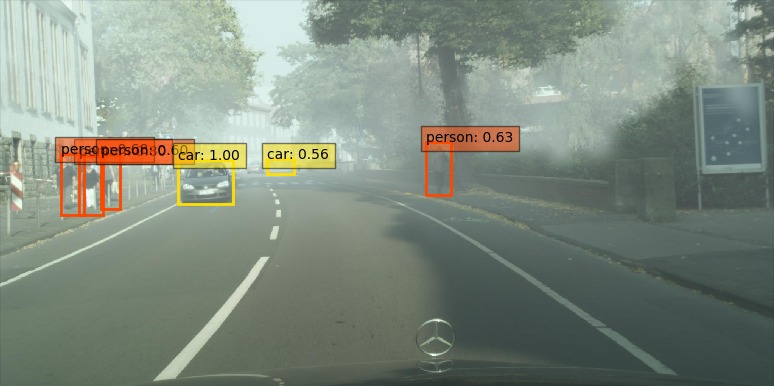}
		\end{subfigure}\\\vspace{0.1cm}
		\begin{subfigure}[t]{0.24\textwidth}\centering
			\includegraphics[width=.98\textwidth]{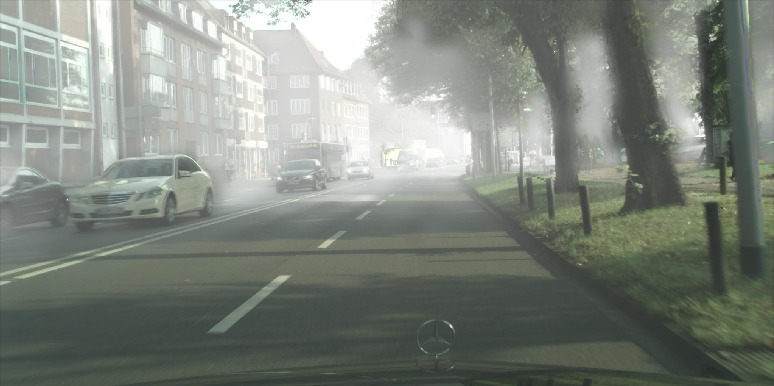}
		\end{subfigure}
		\begin{subfigure}[t]{0.24\textwidth}\centering
			\includegraphics[width=.98\textwidth]{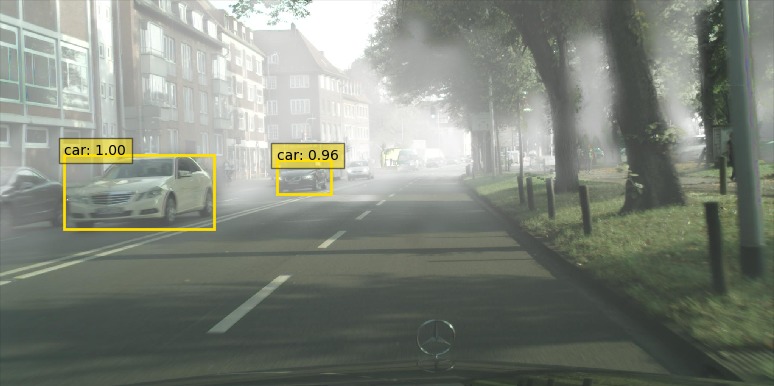}
		\end{subfigure}
		\begin{subfigure}[t]{0.24\textwidth}\centering
			\includegraphics[width=.98\textwidth]{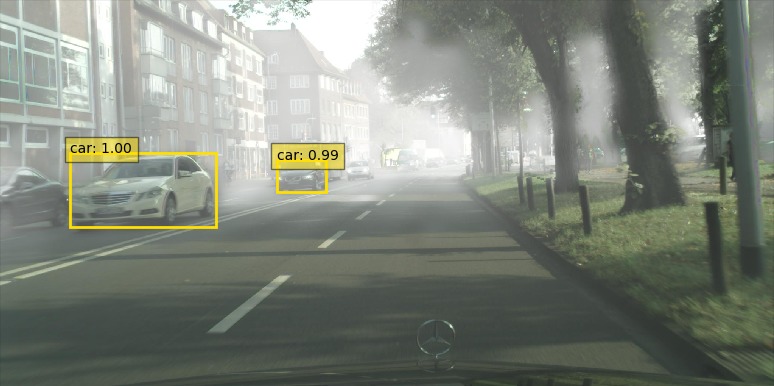}
		\end{subfigure}
		\begin{subfigure}[t]{0.24\textwidth}\centering
			\includegraphics[width=.98\textwidth]{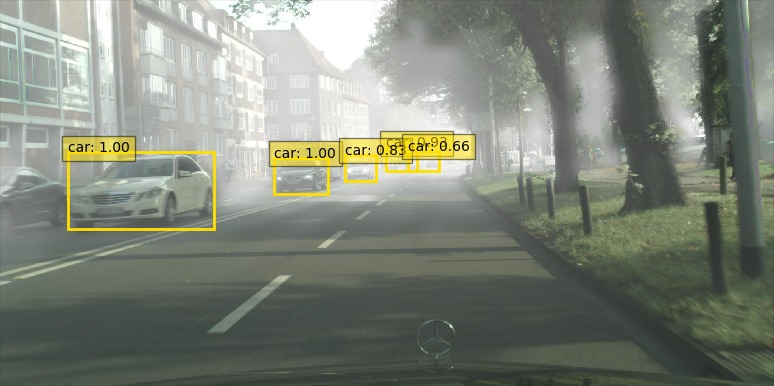}
		\end{subfigure}\\\vspace{0.1cm}
		\begin{subfigure}[t]{0.24\textwidth}\centering
			\includegraphics[width=.98\textwidth]{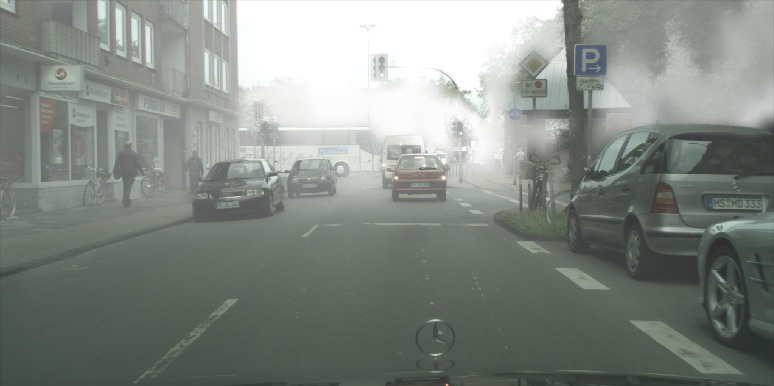}
		\end{subfigure}
		\begin{subfigure}[t]{0.24\textwidth}\centering
			\includegraphics[width=.98\textwidth]{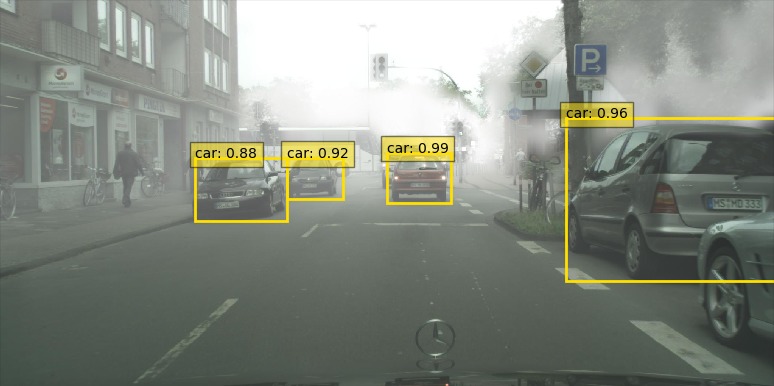}
		\end{subfigure}
		\begin{subfigure}[t]{0.24\textwidth}\centering
			\includegraphics[width=.98\textwidth]{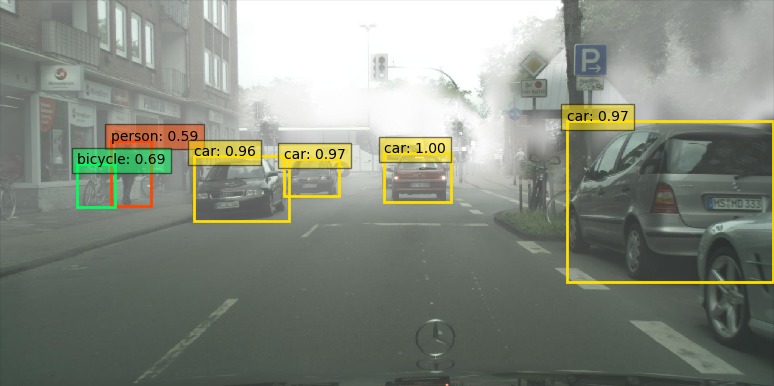}
		\end{subfigure}
		\begin{subfigure}[t]{0.24\textwidth}\centering
			\includegraphics[width=.98\textwidth]{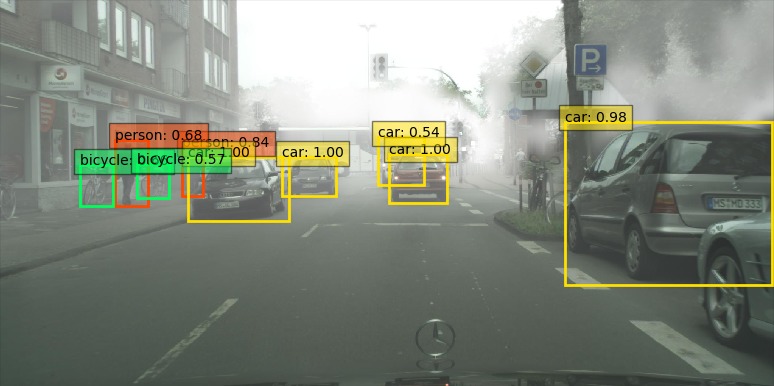}
		\end{subfigure}\\\vspace{0.1cm}
		\begin{subfigure}[t]{0.24\textwidth}\centering
			\includegraphics[width=.98\textwidth]{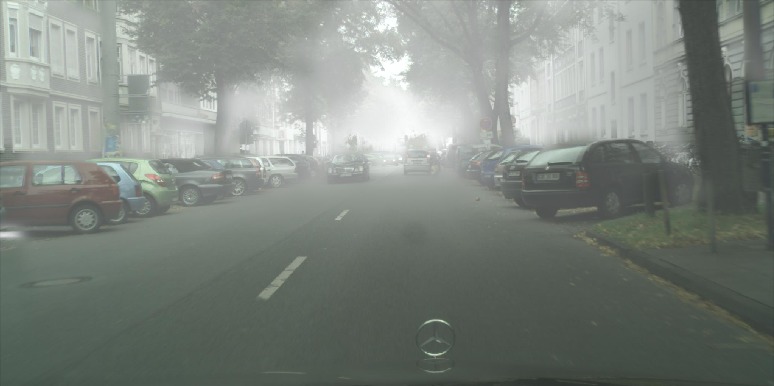}
		\end{subfigure}
		\begin{subfigure}[t]{0.24\textwidth}\centering
			\includegraphics[width=.98\textwidth]{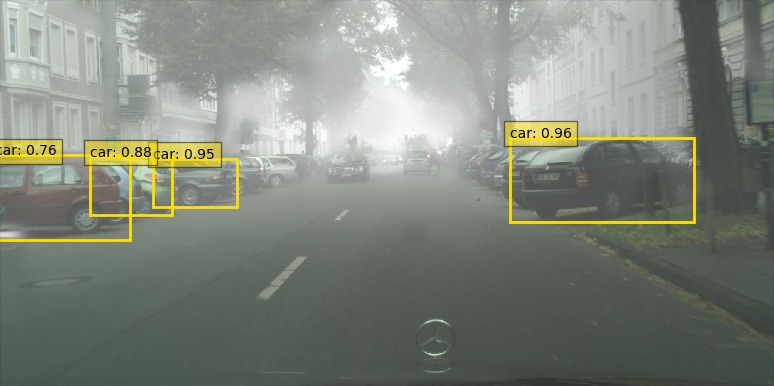}
		\end{subfigure}
		\begin{subfigure}[t]{0.24\textwidth}\centering
			\includegraphics[width=.98\textwidth]{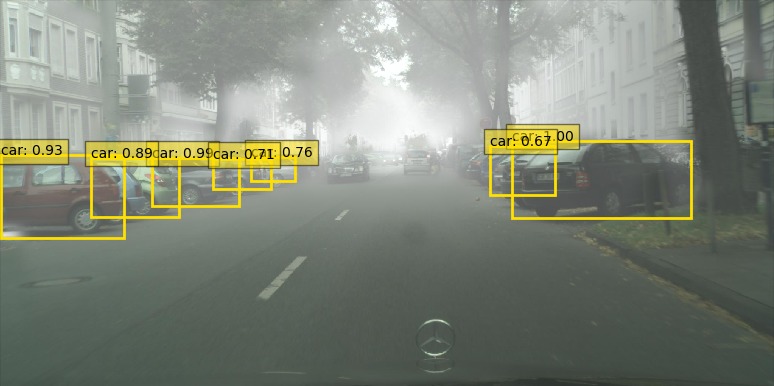}
		\end{subfigure}
		\begin{subfigure}[t]{0.24\textwidth}\centering
			\includegraphics[width=.98\textwidth]{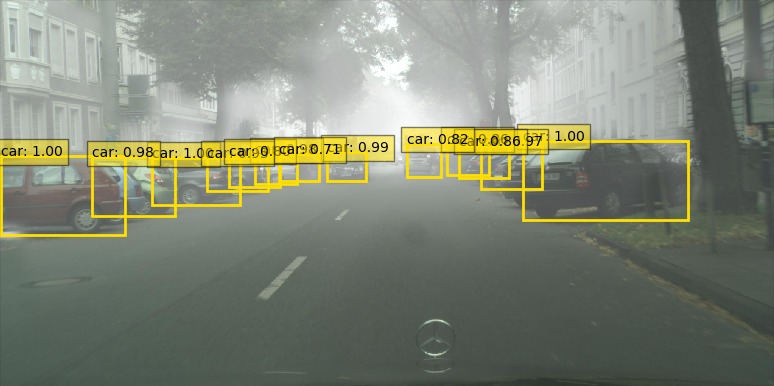}
		\end{subfigure}\\\vspace{0.1cm}
		\begin{subfigure}[t]{0.24\textwidth}\centering
			\includegraphics[width=.98\textwidth]{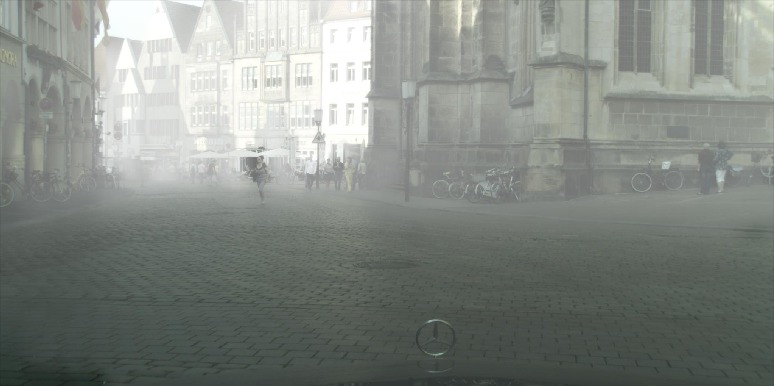}
		\end{subfigure}
		\begin{subfigure}[t]{0.24\textwidth}\centering
			\includegraphics[width=.98\textwidth]{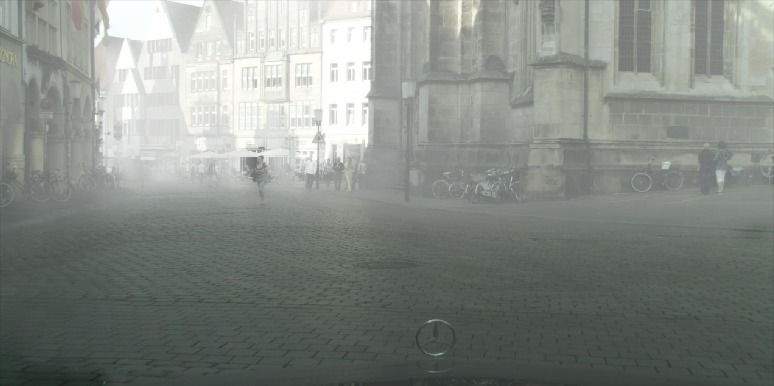}
		\end{subfigure}
		\begin{subfigure}[t]{0.24\textwidth}\centering
			\includegraphics[width=.98\textwidth]{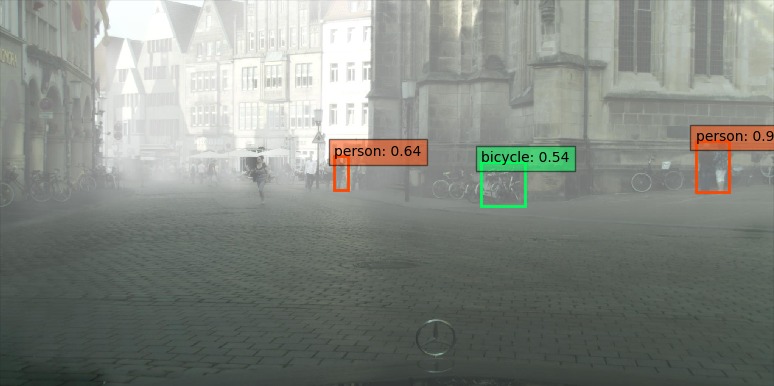}
		\end{subfigure}
		\begin{subfigure}[t]{0.24\textwidth}\centering
			\includegraphics[width=.98\textwidth]{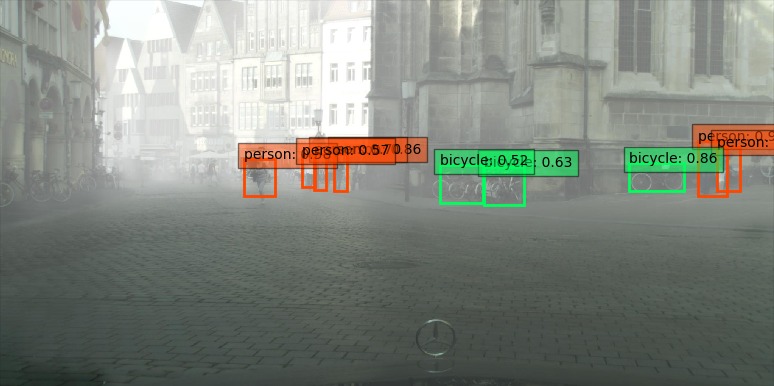}
		\end{subfigure}\\\vspace{0.1cm}
		\begin{subfigure}[t]{0.24\textwidth}\centering
			\includegraphics[width=.98\textwidth]{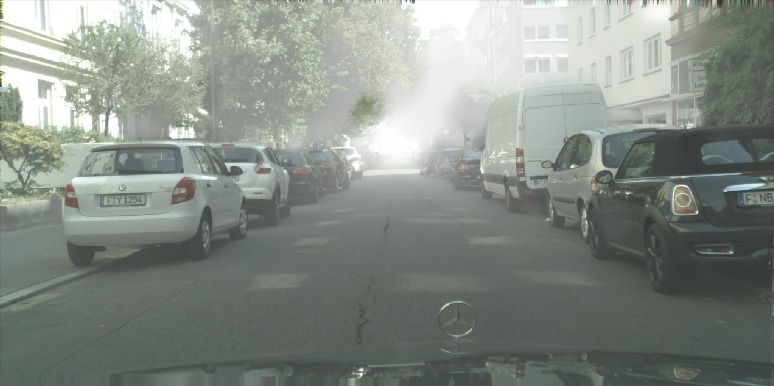}
		\end{subfigure}
		\begin{subfigure}[t]{0.24\textwidth}\centering
			\includegraphics[width=.98\textwidth]{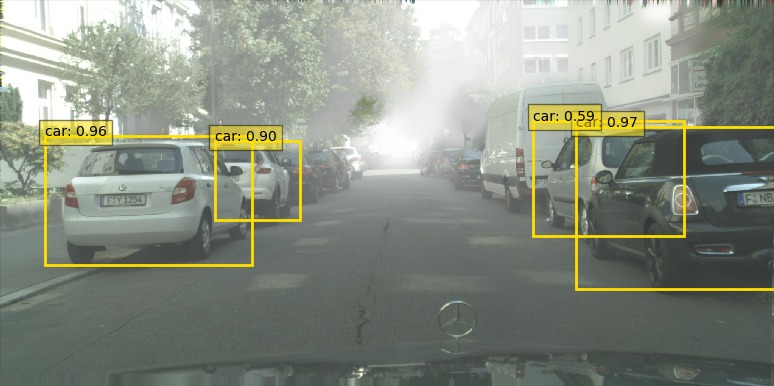}
		\end{subfigure}
		\begin{subfigure}[t]{0.24\textwidth}\centering
			\includegraphics[width=.98\textwidth]{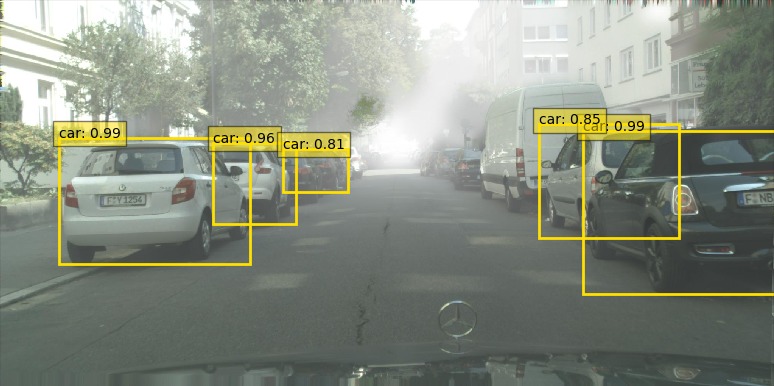}
		\end{subfigure}
		\begin{subfigure}[t]{0.24\textwidth}\centering
			\includegraphics[width=.98\textwidth]{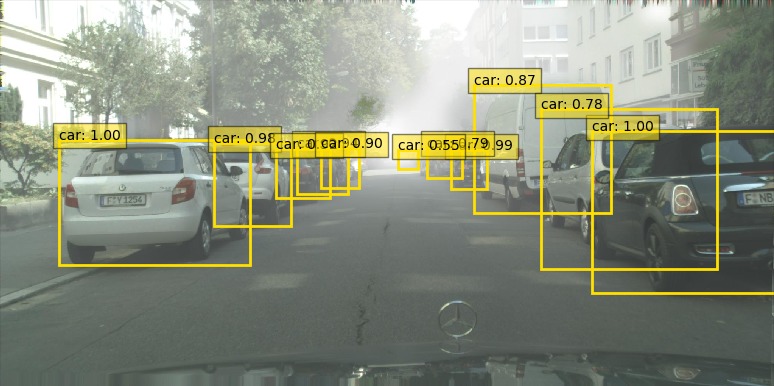}
		\end{subfigure}\\\vspace{0.1cm}
		\begin{subfigure}[t]{0.24\textwidth}\centering
			\includegraphics[width=.98\textwidth]{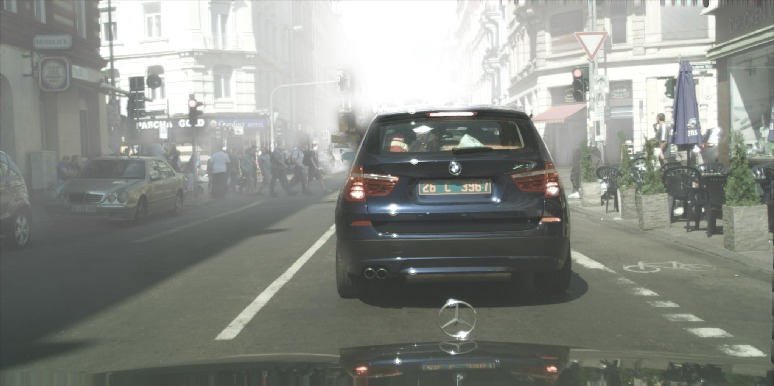}
		\end{subfigure}
		\begin{subfigure}[t]{0.24\textwidth}\centering
			\includegraphics[width=.98\textwidth]{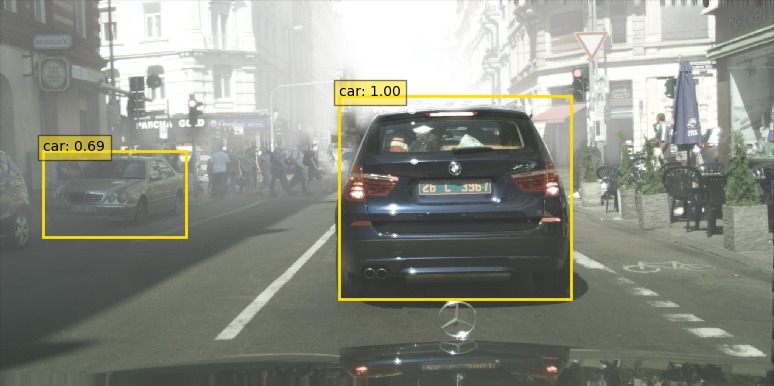}
		\end{subfigure}
		\begin{subfigure}[t]{0.24\textwidth}\centering
			\includegraphics[width=.98\textwidth]{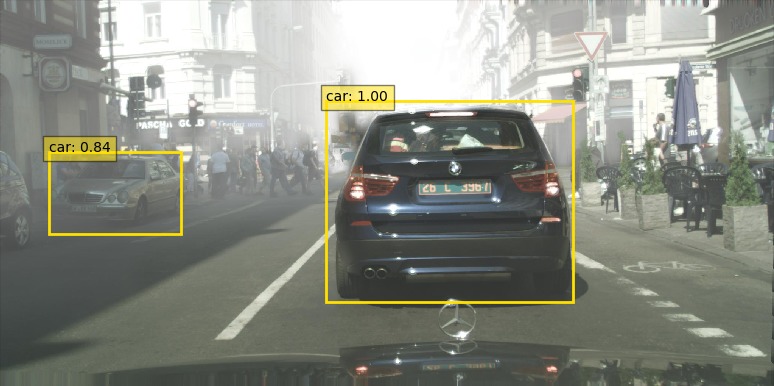}
		\end{subfigure}
		\begin{subfigure}[t]{0.24\textwidth}\centering
			\includegraphics[width=.98\textwidth]{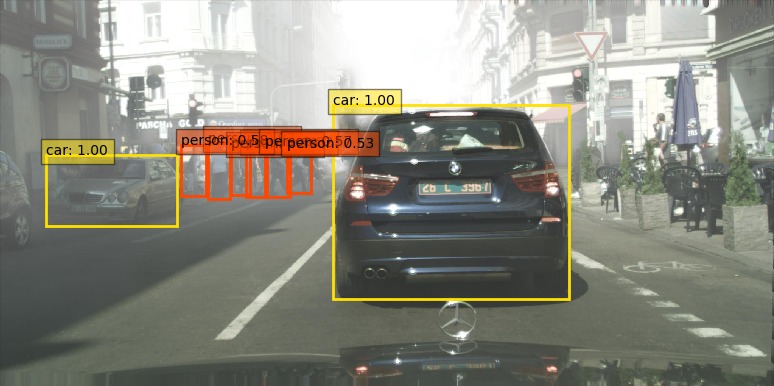}
		\end{subfigure}\\\vspace{0.1cm}
		\begin{subfigure}[t]{0.24\textwidth}\centering
			\includegraphics[width=.98\textwidth]{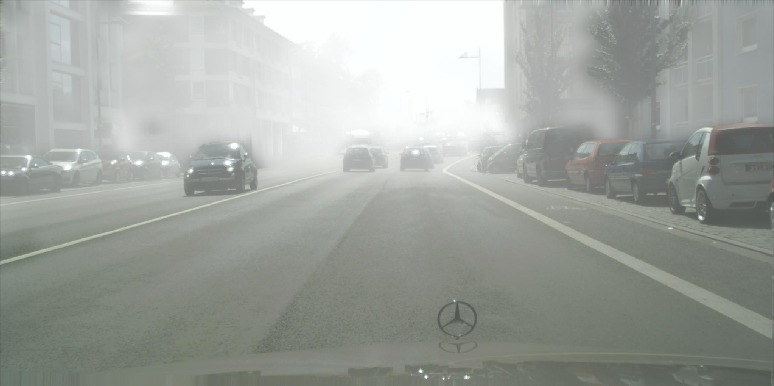}
		\end{subfigure}
		\begin{subfigure}[t]{0.24\textwidth}\centering
			\includegraphics[width=.98\textwidth]{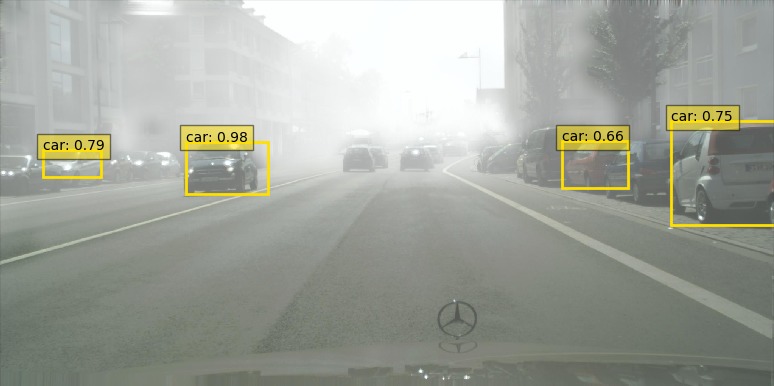}
		\end{subfigure}
		\begin{subfigure}[t]{0.24\textwidth}\centering
			\includegraphics[width=.98\textwidth]{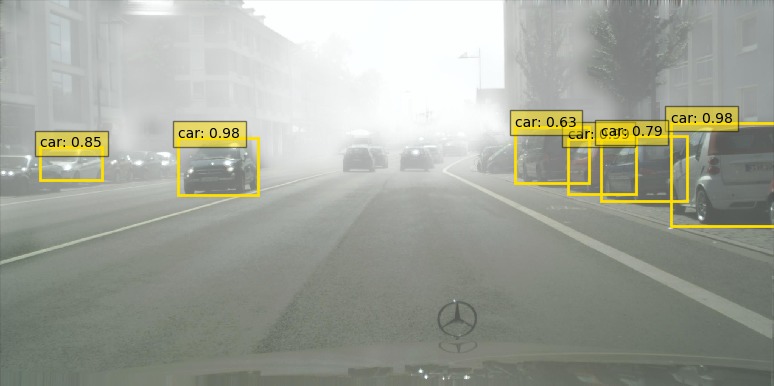}
		\end{subfigure}
		\begin{subfigure}[t]{0.24\textwidth}\centering
			\includegraphics[width=.98\textwidth]{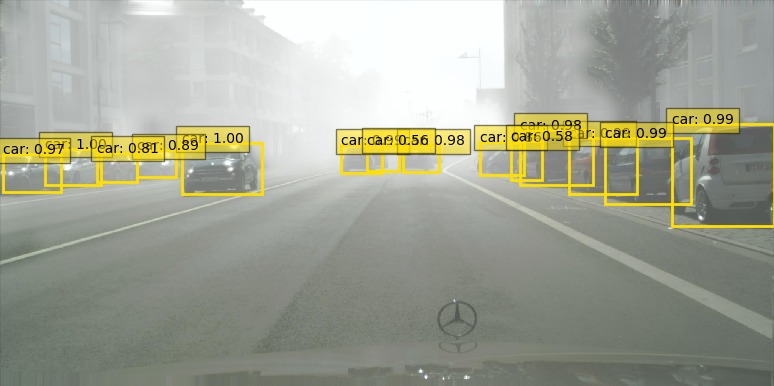}
		\end{subfigure}\\\vspace{0.1cm}
		\begin{subfigure}[t]{0.24\textwidth}\centering
			\includegraphics[width=.98\textwidth]{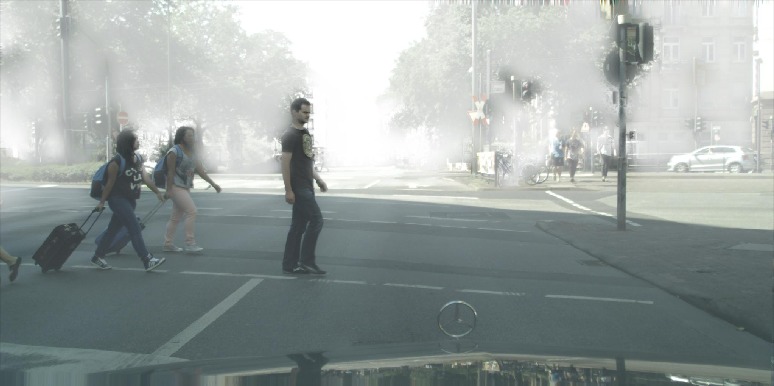}
		\end{subfigure}
		\begin{subfigure}[t]{0.24\textwidth}\centering
			\includegraphics[width=.98\textwidth]{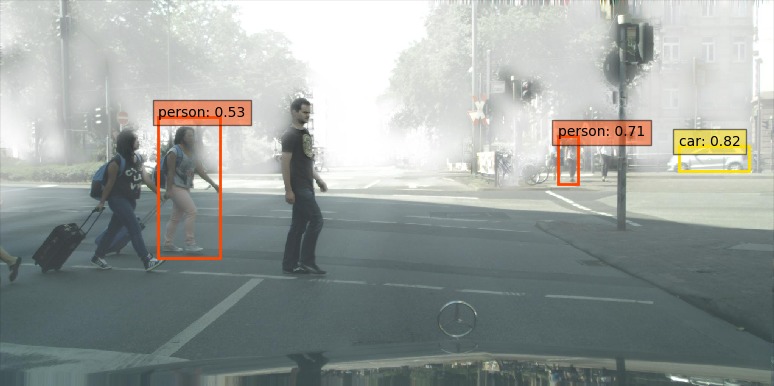}
		\end{subfigure}
		\begin{subfigure}[t]{0.24\textwidth}\centering
			\includegraphics[width=.98\textwidth]{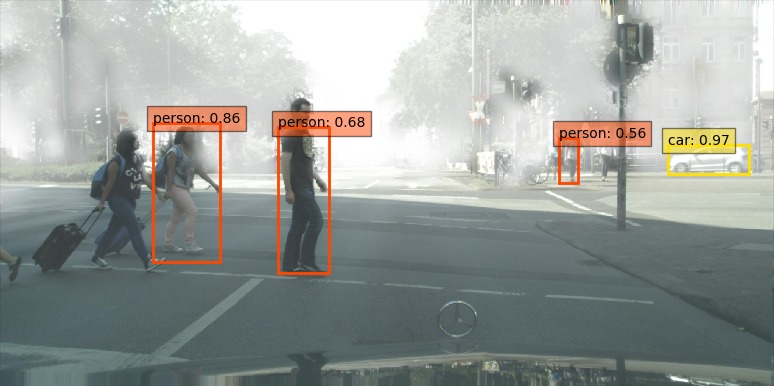}
		\end{subfigure}
		\begin{subfigure}[t]{0.24\textwidth}\centering
			\includegraphics[width=.98\textwidth]{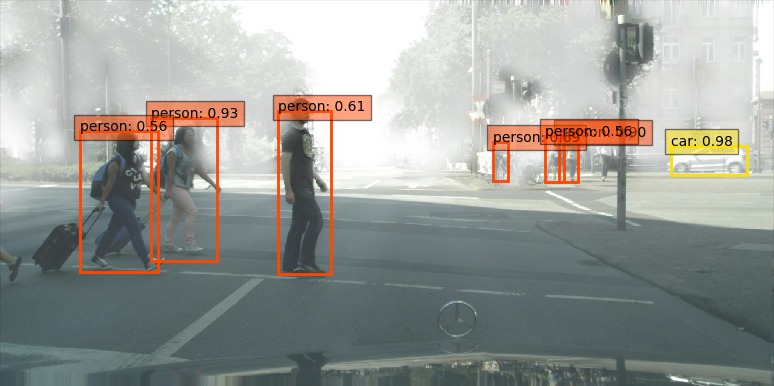}
		\end{subfigure}\\\vspace{0.1cm}
	\end{center}
	\vspace{-0.3cm}
	\caption{\small \textbf{Qualitative results for object detection.} Column (a) shows the input images. Columns (b), (c) and (d) illustrate detection results of the baseline, our MinEnt and AdvEnt models. Detections of different classes are plotted in different colors. We visualize all the detections with scores greater than $0.5$.}
	\vspace{-0.4cm}
	\label{fig:res_ob}
\end{figure*}

\begin{table}[t!]
	\scriptsize{
		\vspace{-0.1cm}
		\begin{center}
			\begin{tabular}{@{}L{2.0cm}@{}|@{}C{0.7cm}@{}C{0.7cm}@{}C{0.7cm}@{}C{0.7cm}@{}C{0.7cm}@{}C{0.7cm}@{}C{0.7cm}@{}C{0.7cm}@{}|@{}C{0.7cm}@{}}		\multicolumn{10}{c}{\small \rule{0pt}{2.5ex} Cityscapes $\rightarrow$ Cityscapes Foggy}\\
				\hline
				\hline
				Models & \rtb{person} & \rtb{rider\,\,} & \rtb{car} & \rtb{truck} & \rtb{bus} & \rtb{train} & \rtb{mcycle} & \rtb{bicycle}  & mAP\\
				\hline
				\rule{0pt}{3ex}SSD-300&15.0&17.4&27.2&5.7&15.1&9.1&11.0&16.7&14.7\\
				\rowcolor[gray]{.92}[0pt][0pt]\rule{0pt}{3ex}Ours (MinEnt)&15.8&22.0&28.3&5.0&15.2&15.0&13.0&20.6&16.9\\
				\rowcolor[gray]{.92}[0pt][0pt]\rule{0pt}{3ex}Ours (AdvEnt)&\textbf{17.6}&\textbf{25.0}&\textbf{39.6}&\textbf{20.0}&\textbf{37.1}&\textbf{25.9}&\textbf{21.3}&\textbf{23.1}&\textbf{26.2}\\
			\end{tabular}
		\end{center}
	}
	\vspace{-0.5cm}
	\caption{\small \textbf{Detection performance on \textit{Cityscapes Foggy}}.
	}
	\vspace{-0.5cm}
	\label{res_ob}
\end{table}
\vspace{-0.45cm}\paragraph{Application on UDA for object detection.}
The proposed entropy-based approaches are not limited to semantic segmentation and can be applied to UDA for other recognition tasks like object detection.
We conducted experiments in the UDA object detection set-up Cityscapes$\rightarrow$Cityscapes-Foggy, similar to the one in~\cite{chen2018domain}.
A straight-forward application of the entropy loss and of the adversarial loss to the existing detection architecture SSD-300~\cite{SSD_2016_ECCV} significantly improves detection performance over the baseline model trained only on source.
In terms of mean-average-precision (mAP), compared to the baseline performance of $14.7\%$, the MinEnt and AdvEnt models attain mAPs of $16.9\%$ and $26.2\%$.
In~\cite{chen2018domain}, the authors report a slightly better performance of $27.6\%$ mAP, using Faster-RCNN~\cite{fasterrcnn}, a more complex detection architecture than ours.
We note that our detection system was trained and tested on images at lower resolution, \ie, $300\times300$.
Despite these unfavorable factors, our improvement to the baseline ($+11.5\%$ mAP using AdvEnt) is larger than the one reported in~\cite{chen2018domain} ($+8.8\%$).
Such a preliminary result suggests the possibility of applying entropy-based approaches on UDA for detection.
Table~\ref{res_ob} reports the per-class IoU and Figure~\ref{fig:res_ob} shows qualitative results of the approaches on \textit{Cityscapes Foggy}.
More technical details are given in the Appendix~\ref{sec:sup_uda_objdet}.

	\section{Conclusion}
	In this work, we address the task of unsupervised domain adaptation for semantic segmentation and propose two complementary entropy-based approaches.
Our models achieve state-of-the-art on the two challenging ``synthetic-2-real'' benchmarks.
The ensemble of the two models improves the performance further.
On UDA for object detection, we show a promising result and believe that the performance can get better using more robust detection architectures.
	
	\appendix
	\section{Entropy-based UDA for object detection}\label{sec:sup_uda_objdet}
	\paragraph{Object detection framework.}
We use the Single Shot MultiBox Detector (SSD-300)~\cite{SSD_2016_ECCV} with the VGG-16 base CNN~\cite{simonyan2014very} as the detection backbone in our experiments.
Given an input image, SSD-300 produces dense predictions from $M$ feature maps at different resolutions.
In detail, every location on each feature map $m$ corresponds to a set of $K_m$ anchor boxes with predefined aspect ratios and scales.
The detection pipeline ends with a non-maximum suppression (NMS) step to post-process the predictions.
Readers are referred to~\cite{SSD_2016_ECCV} for more details about the architecture and the training procedure.
We denote the ``soft-detection map'' of the SSD model at feature map $m$ of dimension $H_m \times W_m$ as $\mm P^m_{\mm x}\in [0,1]^{H_m\times W_m\times K_m\times C}$.

\paragraph{Direct entropy minimization.} Considering a target input image $\trg{\mm x}$, the entropy map produced at a feature map $m$, $\mm E^{m}_{\trg{\mm x}} \in [0,1]^{H_m \times W_m \times K_m}$, is composed of the independent box-level entropies normalized to $[0,1]$:
\begin{equation}
\mm E_{\trg{\mm x}}^{m(h,w,k)} = \frac{-1}{\log(C)}\sum_{c=1}^{C} \mm P_{\trg{\mm x}}^{m(h,w,k,c)} \log \mm P_{\trg{\mm x}}^{m(h,w,k,c)}.
\end{equation}
The entropy loss $\mathcal{L}_{ent}$ is defined as the sum of normalized box entropies over all anchor boxes and all feature resolutions:
\begin{equation}
\mathcal{L}_{ent}(\trg{\mm x}) = \sum_{m}\sum_{h,w,k} \mm E_{\trg{\mm x}}^{m(h,w,k)}.
\end{equation}
Similar to the semantic segmentation task, we jointly optimize the supervised object detection losses on source and the unsupervised entropy loss $\mathcal{L}_{ent}$ on target samples.

\paragraph{Entropy minimization with adversarial learning.}
We apply the adversarial framework proposed in Section~\ref{sec:adv_ent}.
To this end, we first transform the soft-detection maps $P^m_{\mm x}$ to the weighted self-information maps $I^m_{\mm x}$.
We then zero-pad the $I^m_{\mm x}$ maps at lower resolutions to match the size of the largest one.
Finally, we stack all zero-padded $I^m_{\mm x}$ to produce $I_{\mm x}$, which serves as the input to the discriminator.
	
	{\small
		\bibliographystyle{ieee}
		\bibliography{egbib}
	}
\end{document}